\documentclass{article} 
\usepackage{iclr2021_conference,times}

\usepackage{amsmath,amsfonts,bm}

\def\eqref#1{equation~\ref{#1}}

\def\1{\bm{1}}

\DeclareMathAlphabet{\mathsfit}{\encodingdefault}{\sfdefault}{m}{sl}
\SetMathAlphabet{\mathsfit}{bold}{\encodingdefault}{\sfdefault}{bx}{n}

\usepackage{url}
\usepackage{amssymb}
\usepackage{lineno}
\usepackage{tikz}
\usepackage{caption}
\usepackage{float}
\usepackage{pgfplots}
\usepackage{subfig}
\usepackage[utf8]{inputenc}
\usepackage{graphicx}
\usepackage{float}
\usepackage{listings}
\usepackage{verbatim}
\usepackage[shortcuts,acronym]{glossaries}
\usepackage{color}
\usepackage{comment}
\usepackage{hyperref}
\usepackage[utf8]{inputenc}

\usepackage[acronym]{glossaries}
\makeglossaries

\definecolor{mycolor1}{rgb}{0.00000,0.44700,0.74100}
\definecolor{mycolor2}{rgb}{0.85000,0.32500,0.09800}
\definecolor{mycolor3}{rgb}{0.92900,0.69400,0.12500}

\usepackage{enumitem}
\usepackage[utf8]{inputenc} 
\usepackage[T1]{fontenc}    

\usepackage{url}            
\usepackage{booktabs}       
\usepackage{amsfonts}       
\usepackage{nicefrac}       
\usepackage{microtype}      
\usepackage{tikz}

\usepackage{xcolor}
\usepackage[acronym]{glossaries}

\makeglossaries
\usepackage{cleveref}
\usepackage{multirow}
\usepackage{color}

\usepackage{wrapfig,lipsum}
\usepackage{subfig}

\usepackage{amsthm}

\newacronym{MNIST}{MNIST}{Modified National Institute of Standards and Technology dataset}
\newacronym{SVHN}{SVHN}{Street View House Numbers}
\newacronym{CIFAR-10}{CIFAR-10}{Canadian Institute for Advanced Research dataset of 10 classes}
\newacronym{IID}{IID}{independently and identically distributed}
\newacronym{TI}{TI}{Tiny ImageNet}
\newacronym{SAPN}{SAPN}{Salt and Pepper Noise}
\newacronym{FP}{FP}{Fashion Product}
\newacronym{GN}{GN}{Gaussian Noise}

\usepackage{comment}

\newacronym{OOD}{OOD}{Out of Distribution}
\newacronym{IOD}{IOD}{Inside of Distribution}
\newacronym{DeDiM}{DeDiM}{Deep data set Dissimilarity Measure}

\newacronym{SSDL}{SSDL}{semi-supervised deep learning}

\newacronym{OH}{OH}{Other-Half}
\newacronym{Sim}{Sim}{Similar}
\newacronym{Dif}{Dif}{Different}

\title{More Than Meets The Eye: Semi-supervised Learning Under Non-IID Data}

\author{Saul Calderon-Ramirez\thanks{Equal contribution}\\
Centre for Computational Intelligence (CCI)\\
De Montfort University\\
Leicester, United Kingdom\\
\texttt{sacalderon@itcr.ac.cr} \\
\And
Luis Oala$^{*}$ \\
AI Department \\
Fraunhofer HHI \\
Berlin, Germany \\
\texttt{luis.oala@hhi.fraunhofer.de} \\
}

\iclrfinalcopy 
\begin{document}

\maketitle

\begin{abstract}
A common heuristic in semi-supervised deep learning (SSDL) is to select unlabelled data based on a notion of semantic similarity to the labelled data. For example, labelled images of numbers should be paired with unlabelled images of numbers instead of, say, unlabelled images of cars. We refer to this practice as semantic data set matching. In this work, we demonstrate the limits of semantic data set matching. We show that it can sometimes even degrade the performance for a state of the art SSDL algorithm. We present and make available a comprehensive simulation sandbox, called \textit{non-IID-SSDL}, for stress testing an SSDL algorithm under different degrees of distribution mismatch between the labelled and unlabelled data sets. In addition, we demonstrate that simple density based dissimilarity measures in the  feature space of a generic classifier offer a promising and more reliable quantitative matching criterion to select unlabelled data before SSDL training.
\end{abstract}

\maketitle

Training an effective deep learning solution typically requires a considerable amount of labelled data $S_l$. In specific areas, like medical imaging, annotations can be expensive to obtain.
Several approaches have been developed to address this data constraint, including data augmentation, self-supervised and \gls{SSDL}, among others \citep{weiss2016survey,wang2017effectiveness,zhou2018brief}. Semi-supervised learning leverages unlabelled data $S_u$, thus offering a promising approach for problems where little labelled data is available or a range of labels is lacking \citep{van2020survey}. 
As with other learning paradigms, the transfer of \gls{SSDL} techniques from lab to real-world is complicated by violations of the \gls{IID} assumption, 
specifically where the distributions of labelled inputs $P_l$ and unlabelled inputs $P_u$ do not match up, i.e. $P_l \neq P_u$. 
In principle, we would like to exploit available unlabelled data as flexibly as possible. In practice, mismatches between the labelled and unlabelled data sets can lead to serious performance degradation \citep{oliver2018realistic}.
This begs the question how we can systematically select labelled and unlabelled data in non-\gls{IID} settings such that performance on the downstream task is increased. 
A common recourse are what we call semantic matching heuristics. Under such a scheme 
anthropogenic semantic concepts are used to delineate distribution mismatches between data sets. For example, different classes of the same data set have been viewed as out-of-distribution to one another because one class displays animals and the other class displays vehicles \citep{chen2020semi,zhao2021robust}.
Practices of semantic matching can be traced to other fields of machine learning, too, including out-of-distribution detection \citep{zisselman2020deep} or the domain adaptation \citep{zhang2020label} literature. 
Insights from generative modelling hint at the limits of semantic matching: a supposedly different data set like \gls{SVHN} can have a higher likelihood under a generative model trained on ImageNet than ImageNet data itself \citep{nalisnick2018deep}. 
Thus, in this work we take a close look at \gls{SSDL} under varying non-IID conditions and demonstrate the limits of semantic matching. Specifically:
\begin{itemize}[leftmargin=*]
    \item We show that semantic heuristics can fail when selecting labelled and unlabelled data for the state-of-the-art \gls{SSDL} algorithm MixMatch. To our knowledge, this is the first work to extensively evaluate the accuracy behaviour of MixMatch under different degrees of mismatch between the labelled and unlabelled data. We make available the comprehensive simulation sandbox, called \textit{non-\gls{IID}-\gls{SSDL}}, from our experiments. The sandbox incorporates eight data sets and different configurations for mixing data sets and the number of labelled data points.
    \item In addition, we demonstrate that simple dissimilarity measures in the feature space of a generic deep ImageNet classifier offer a promising and more reliable matching criterion. We corroborate these results with a comprehensive statistical analysis.
    
\end{itemize}
\section{Method}
Note that from hereon out we refer to \gls{IOD} data as the labelled, in-distribution data and to \gls{OOD} data as the unlabelled data that is non-\gls{IID} relative to \gls{IOD}. 
In order to systematically assess the impact of non-\gls{IID} data on \gls{SSDL} performance we created the flexible \textit{non-\gls{IID}-\gls{SSDL}} simulation sandbox which we make openly available\footnote{All code and experimental scripts, with automatic download of sandbox data for ease of reproduction, can be found at \url{https://github.com/luisoala/non-iid-ssdl}.}.
The ablation sandbox has five degress of freedom: 
base data $S_{\textrm{IOD}}$ which constitutes the original task to be learned, the type of \gls{OOD} data $T_{\textrm{OOD}}$, the \gls{OOD} data source $S_{\textrm{u,OOD}}$, the relative amount of \gls{OOD} data among the unlabelled data $\%_{\textrm{u,OOD}}$, the amount $n_l$ of labelled observations and the \gls{SSDL} algorithm to be used. More details on the configurations of these arguments are provided in \Cref{sec:experiments} below. The algorithm under consideration for the \textit{non-\gls{IID}-\gls{SSDL}} stress test in our analysis is MixMatch, a state of the art \gls{SSDL} method \citep{berthelot2019mixmatch}. We provide a detailed description of our implementation in \Cref{app:algo,app:hyper}.
As a simple quantitative baseline against the semantic matching heuristic we consider the dissimilarity between two data sets $S_a$ and $S_b$ in the feature space of a generic Wide-ResNet pre-trained on ImageNet. 
We consider four dissimilarity measures: two Minkowski based distance sets, $d_{\ell_{2}}\left(S_{a},S_{b},\tau, \mathcal{C}\right)$ and $d_{\ell_{1}}\left(S_{a},S_{b},\tau, \mathcal{C}\right)$, corresponding to the Euclidean and Manhattan distances, respectively.
Full details on the computation of the measures can be found in \Cref{app:measures}. We also perform additional qualitative analyses by comparing the distributions of individual features between different data sets.

Finally, we corroborate the results with statistical analyses. The distances were all calculated on smaller sub-samples of the full data sets. Hence, we also test if the measurements are statistically meaningful using a Wilcoxon test \citep{fagerland2009wilcoxon}.  In all sandbox simulations, we performed ten experimental runs and report the accuracy mean and variance of the models performing best on the test data from each run.
\section{Experiments}
\label{sec:experiments}
\begin{table*}[]
\vspace{-0.5cm}
\caption{Results for the class distribution mismatch experiment, best OOD performance in bold per configuration. Each result entry in the table represents the mean and variance of accuracy across ten random experimental runs per entry. For a detailed description of symbols and the experiment see \Cref{sec:experiments}. Numbers in parentheses next to $S_\textrm{u,OOD}$ entries reflect the preference ranking provided by disimilarity matching with $d_{C}$.}
\label{tab:selected-results}

\centering

\scalebox{0.5}{

\begin{tabular}{ccccccc}
\toprule
\multirow{2}{*}{$\mathbf{S_{IOD}}$} & \multirow{2}{*}{$\mathbf{T_{OOD}}$} & \multirow{2}{*}{$\mathbf{S_\textrm{u,OOD}}$} & \multirow{2}{*}{$\mathbf{\%_{uOOD}}$} & \multicolumn{3}{c}{$\mathbf{n_l}$}\\

&  & &  & 60 & 100 & 150\\

\midrule

\multirow{12}{*}{\rotatebox[origin=c]{90}{\textbf{MNIST}}} 
& \multicolumn{3}{c}{Fully supervised baseline} & $0.457\pm0.108$ & $0.559\pm 0.125 $ & $0.645 \pm0.101$ \\

 & \multicolumn{3}{c}{SSDL baseline (no OOD data)} & $0.704 \pm 0.096$ & $0.781\pm 0.065$ & $0.831 \pm0.0626$\\

 \cline{2-7}
 
 & \multirow{2}{*}{OH} & \multirow{2}{*}{MNIST} & 50 & $\mathbf{0.679\pm0.108}$ & $\mathbf{0.769\pm0.066}$ & $0.802\pm0.054$\\
 & & & 100 & $0.642 \pm 0.111$ & $0.746 \pm0.094$ & $0.798 \pm0.07$\\

 \cline{2-7}
 & \multirow{2}{*}{Sim} & \multirow{2}{*}{SVHN (3)} & 50 & $0.637 \pm0.098$ & $0.745 \pm0.081$ & $0.801 \pm0.0699$\\
 
 & & & 100 & $0.482 \pm0.113$ & $0.719 \pm0.058$ & $0.765 \pm0.072$\\
 
  \cline{2-7}
 
  & \multirow{6}{*}{Dif}& \multirow{2}{*}{TI (2)} & 50 & $0.642 \pm0.094$ & $0.739 \pm 0.074$ & $0.809 \pm0.066$\\
  
  & & & 100 & $0.637 \pm0.097$ & $0.732\pm 0.074$ & $0.804 \pm0.071$\\

 \cline{3-7}
 
  & & \multirow{2}{*}{GN (5)} & 50 & $0.606 \pm0.0989$ & $0.713 \pm0.087$ & $0.786\pm 0.065$\\
  
  & & & 100 & $0.442 \pm0.099$ & $0.461 \pm0.073$ & $0.542\pm 0.062$\\

   \cline{3-7}
 
  & & \multirow{2}{*}{SAPN (4)} & 50 & $0.631 \pm0.102$ & $0.735 \pm0.082$ & $\mathbf{0.813 \pm0.057}$\\
  
  & & & 100 & $0.48\pm0.0951$ & $0.524 \pm0.09$ & $0.563 \pm0.095$\\
 \bottomrule
\end{tabular}

}
\vspace{-.5cm}
\end{table*}
In order to evaluate semantic matching heuristic we consider the following settings for the \textit{non-\gls{IID}-\gls{SSDL}} sandbox. Three configurations for $S_{\textrm{IOD}}$ comprising MNIST, CIFAR-10 and FashionMNIST. A total of three configurations for $T_{\textrm{OOD}}$ (\gls{OH}, \gls{Sim} and \gls{Dif}) are tested. In the \gls{OH} setting half of the classes and associated inputs are taken to be the $S_{\textrm{IOD}}$ data whereas the other half of classes are taken to be the $S_{\textrm{u,OOD}}$ data. \gls{Sim} is a $S_{\textrm{u,OOD}}$ data set that is supposedly semantically related to $S_{\textrm{IOD}}$, e.g. MNIST and \gls{SVHN}. \gls{Dif} is  a $S_{\textrm{u,OOD}}$ data set that is assumed to be semantically unrelated to $S_{\textrm{IOD}}$, e.g. MNIST and Tiny ImageNet. We note that a degree of subjectivity is involved in the selection what is to be considered \gls{Sim} and \gls{Dif} - a drawback of the semantic matching heuristic itself. There are five configurations for $S_{\textrm{u,OOD}}$ as explained above: the other half \gls{OH}, a similar data set, and three different data sets including two noise baselines. They include Street View House Numbers \gls{SVHN}, \gls{TI}, \gls{GN}, \gls{SAPN} and \gls{FP}.  Each configuration represents a multi-class classification task with $|\mathcal{Y}|=5$, that is a random subset of half of the classes of base data $S_{\textrm{IOD}}$. This is so that metrics on all $T_{\textrm{OOD}}$ settings, including \gls{OH}, are comparable. We vary the relative amount of \gls{OOD} data $\%_{\textrm{u,OOD}}$ between 0, 50 and 100, where the \gls{IOD} data is a partition of the labelled data. For instance in MNIST, after splitting the classes for \gls{IOD} and \gls{OOD}, we use a portion of the \gls{IOD} data as labelled and another one as unlabelled. We also vary the amount of labelled data points $n_l$ between 60, 100 and 150. We study the behaviour of MixMatch under very limited number of labels settings, where the benefit of \gls{SSDL} is usually higher. This makes the impact of distribution mismatch more evident. For each run we sampled  non-overlapping subsets of data from $S_{\textrm{IOD}}$ and $S_{\textrm{u,OOD}}$ to obtain the required number of labelled $n_l$ and unlabelled $n_u$ samples for the run. Descriptive statistics for standardization of the inputs were only computed from these subsets to keep the simulation realistic and not leak any information from the global data sets. All hyper-parameters are described in Table \ref{table:globalhyper} in \Cref{app:algo} and kept constant across all the tests performed to isolate the effects of the variable sandbox parameters.

Table  \ref{tab:selected-results} shows the results of the data set matching experiments described in \Cref{sec:experiments}. We make a number of observations. 
First, in the majority of cases using \gls{IOD} unlabelled data or a 50-50 mix of \gls{IOD} and \gls{OOD} unlabelled data beats the fully supervised baseline. The gains range from 15\% to 25\% for MNIST, 10\% to 15\% for CIFAR-10 and 7\% to 13\% for FashionMNIST across all $S_{\textrm{u,OOD}}$ and $n_l$. As expected, in most of the cases the accuracy is degraded when including \gls{OOD} data in $S_u$, with a more severe degradation when noisy data sets (\gls{SAPN}, \gls{GN}) are used as  \gls{OOD} data contamination source.
Second, it is not always the case that $T_{\textrm{OOD}}=\textrm{OH}$ (when $S_{\textrm{u,OOD}}$ is supposedly more semantically  similar to $S_{\textrm{IOD}}$) yields the best MixMatch performance. This is observed when $S_l = $ CIFAR-10,  $n_l=100$ and $n_l=150$, where \gls{OOD} unlabelled data from Tiny ImageNet results in more accurate models than using the other half of CIFAR-10 as $S_{\textrm{u,OOD}}$. It is interesting that an $S_{\textrm{u,OOD}}$ data set of type \gls{Dif} can be more beneficial than a $S_{\textrm{u,OOD}}$ data set of type \gls{Sim} which is also the case for FashionMNIST and Tiny ImageNet versus Fashion Product at $n_l=150$. This contradicts the belief underlying the semantic matching heuristic that unlabelled data that appears semantically more related to the labelled data is always the better choice for \gls{SSDL}. Given length restrictions, the full results can be found in \Cref{tab:half-half-results} of \Cref{app:exp-sandbox}.

\begin{table*}[]
\vspace{-.5cm}
\caption{Dissimilarity measures between the labelled and unlabelled datasets $S_l$ and $S_u$. Numbers in italics correspond to results with $p>0.05$ for the Wilcoxon test. For a detailed description of symbols and the experiment see \Cref{app:measures}}
\label{tab:distances-selected}
\begin{center}
\scalebox{0.5}{
\begin{tabular}{ccccccc}
\toprule
$\mathbf{S_l}$     & $\mathbf{S_u}$  & $\mathbf{\%_{uOOD}}$  & $\mathbf{d_{\ell_2}}$  & $\mathbf{d_{\ell_1}}$  & $\mathbf{d_{JS}}$                      & $\mathbf{d_{C}}$                        \\ 
\midrule
\multirow{10}{*}{\rotatebox[origin=c]{90}{\textbf{MNIST}}}           & \multirow{2}{*}{OH}          & 50                   & $\mathit{0.011 \pm0.006}$         & $\mathit{0.459 \pm0.28}$          & $\mathit{0.266 \pm0.221}$                        & $\mathit{0.811 \pm0.512}$                         \\    
& & 100                   & $\mathit{0.014\pm0.019}$        & $\mathit{0.38\pm0.507}$         & $1.001 \pm0.725$                       & $1.263\pm0.665$                         \\ 
\cline{2-7}
                            & \multirow{2}{*}{\gls{SVHN}}            & 50                    & $\mathit{0.09\pm0.017}$         & $1.569\pm0.504$        & $6.789 \pm0.924$                       & $12.021\pm1.757$                        \\
                            &                 & 100                   & $0.25\pm0.053$         & $4.702 \pm 1.04$       & $52.349 \pm 3.292$                     & $42.026\pm4.31$                         \\ 
\cline{2-7}
                            & \multirow{2}{*}{\gls{TI}}              & 50                    & $0.008\pm0.023$       & $1.519\pm0.223$        & $3.663 \pm0.772$                       & $5.512 \pm0.767$                        \\
                            &                 & 100                   & $0.217\pm 0.04$        & $4.3 \pm 0.636$        & $10.305 \pm1.667$                      & $15.18\pm2.698$                         \\ 
\cline{2-7}
                            & \multirow{2}{*}{\gls{GN}}               & 50                    & $0.11 \pm 0.0219$      & $1.958\pm0.534$        & $14.785 \pm1.052$                      & $23.593\pm1.859$                        \\
                            &                 & 100                   & $0.357\pm0.081$        & $5.987 \pm 1.091$      & $52.349\pm4.253$                       & ~$86.21\pm3.471$~                       \\ 
\cline{2-7}
                            & \multirow{2}{*}{\gls{SAPN}}             & 50                    & $0.089\pm0.0311$       & $2.479\pm0.7433$       & $15.116 \pm1.475$                      & $20.151 \pm1.619$                        \\
                            &                 & 100                   & $0.323 \pm 0.07$       & $6.308 \pm 1.366$      & ~$53.397 \pm4.253$                     & $77.456\pm4.474$                       \\ 
\bottomrule
\end{tabular}
}

\vspace{-.5cm}
\end{center}
\end{table*}

\begin{figure}
    \vspace{-1cm}
    \centering
 
    \subfloat[][]{
        \label{tab:correlations}
        \scalebox{0.5}{

        \begin{tabular}{cccccc}
        \toprule
        \textbf{$\mathbf{S_l}$}       & \textbf{$\mathbf{n_l}$} & \textbf{$\mathbf{d_{\ell_1}}$} & \textbf{$\mathbf{d_{\ell_2}}$} & \textbf{$\mathbf{d_{JS}}$} & \textbf{$\mathbf{d_{C}}$} \\ \hline
        \multirow{3}{*}{MNIST}        & 60                      & -0.876                         & -0.898                         & -0.969                     & -0.944                    \\
                                      & 100                     & -0.805                         & -0.83                          & -0.786                     & -0.948                    \\
                                      & 150                     & -0.794                         & -0.822                         & -0.81                      & -0.944                    \\ \hline
        \multirow{3}{*}{CIFAR-10}     & 60                      & -0.823                         & -0.853                         & -0.944                     & -0.921                    \\
                                      & 100                     & -0.826                         & -0.878                         & -0.966                     & -0.947                    \\
                                      & 150                     & -0.808                         & -0.838                         & -0.952                     & -0.927                    \\ \hline
        \multirow{3}{*}{FashionMNIST} & 60                      & -0.2                           & -0.268                         & -0.735                     & -0.789                    \\
                                      & 100                     & -0.264                         & -0.326                         & -0.781                     & -0.824                    \\
                                      & 150                     & -0.286                         & -0.347                         & -0.785                     & -0.827                    \\  
        \bottomrule
        \vspace{1cm}
        \end{tabular}
        }
    }
    \subfloat[][]{
        \begin{tabular}{cc}
            \scalebox{0.24}{\includegraphics{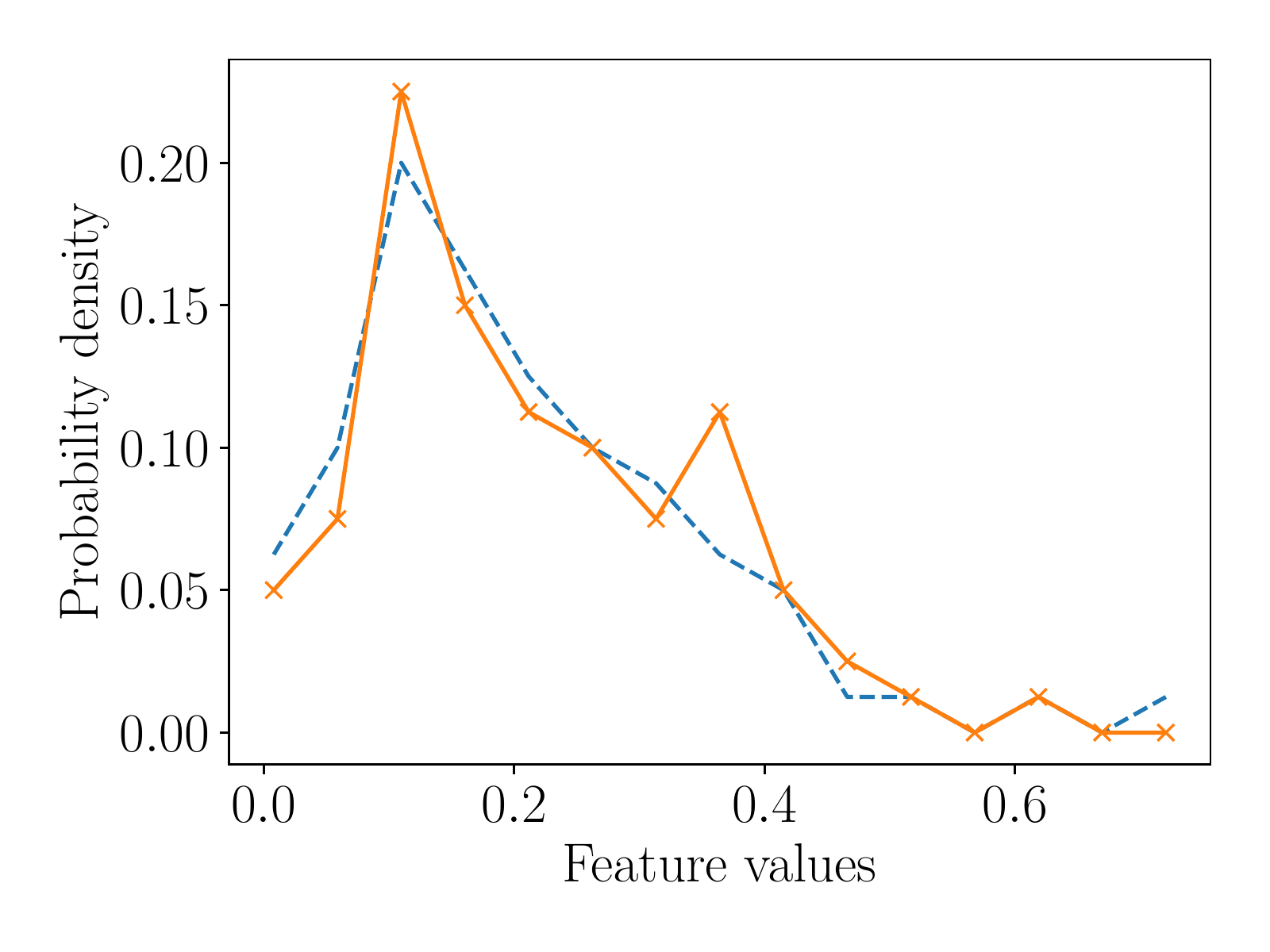}} & \scalebox{0.24}{\includegraphics{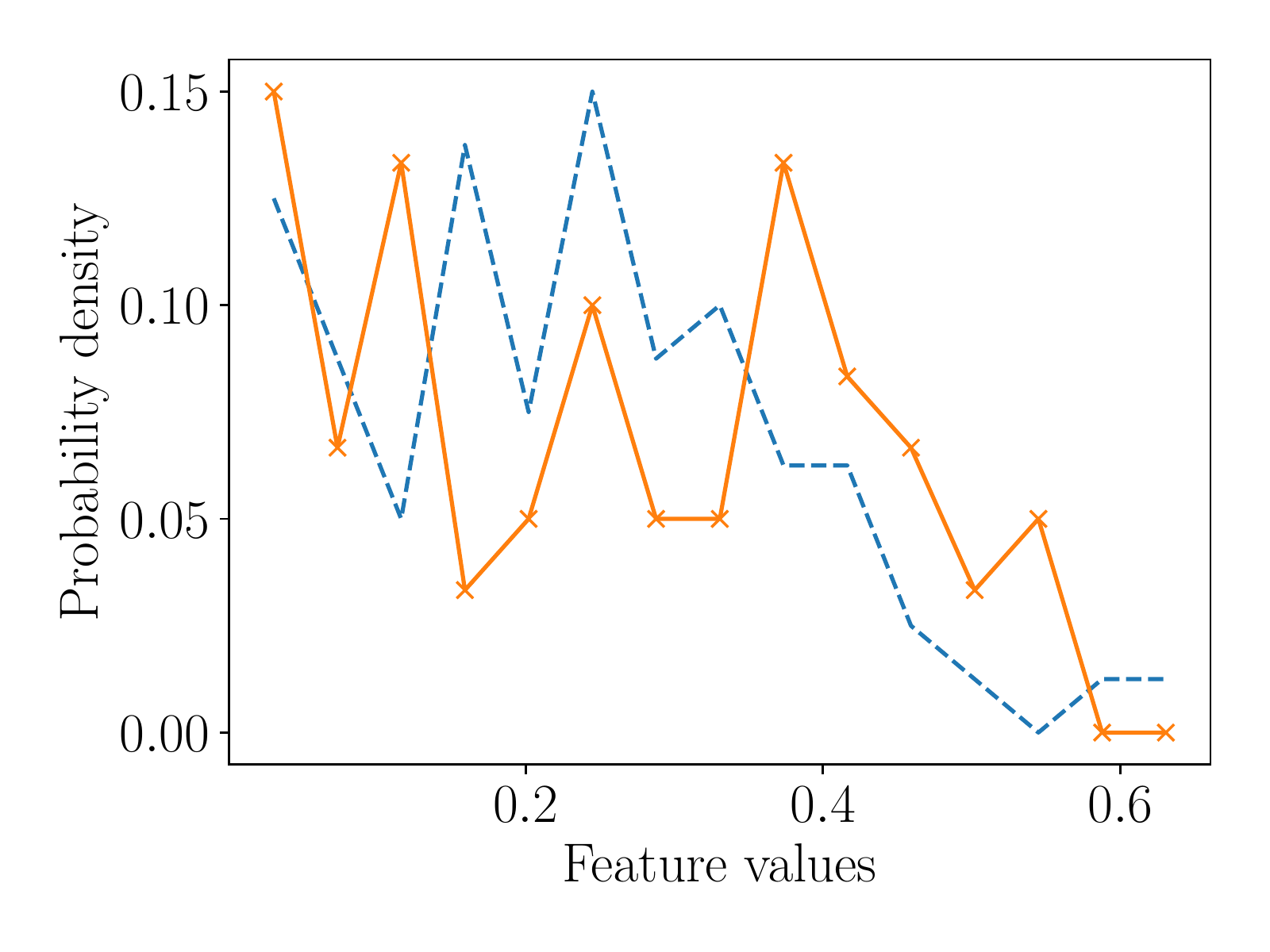}}
        \end{tabular}
        \label{tab:Feature_plots-selected}
    }
       \caption{A summary of the data set dissimilarity results for the test  bed. (a) Correlation results for the dissimilarity measures between $S_l$ and $S_u$ with  \gls{OOD} contamination and  \gls{SSDL} accuracy. (b) Distribution for feature 159 of a model trained with MNIST labelled data (orange and continuos in both plots), and ImageNet and \gls{SVHN} unlabelled data (left and right column, respectively, with the blue dashed line in both).}
\vspace{-.5cm}

\end{figure}

Rather, as we demonstrate in the second set of experiments below, a notion of dissimilarity between data sets in the feature space of a generic deep classifier can offer a simple and reproducible comparison heuristic.
The disimilarity scores for all \gls{OOD} configurations from the ablation study can be found in \Cref{tab:distances-selected}. We can observe that these dissimilarity measures trace the accuracy results found in \Cref{tab:half-half-results}. This correlation is quantified in \Cref{tab:correlations} with the cosine based density measure $d_c$  correlating particularly well with the accuracy results of \Cref{tab:half-half-results}. Also, the p-values  are consistently lower for the density based distances, meaning that density based distances present more confidence as seen in \Cref{tab:distances}. The results with italics correspond to p-values lower than 0.05. We suspect that this is related to the quantitative approximation of the feature distribution mismatch implemented both in the $d_{\textrm{JS}}$ and $d_{\textrm{C}}$ distances. In \Cref{tab:half-half-results} we indicate the distance-based preference ranking provided by disimilarity matching with $d_{C}$. The non-IID configurations resulting in the best \gls{SSDL} accuracy are contained in the top two $d_{C}$ selections seven out of nine times. Note that with the dissimilarity measures we can do this selection \emph{before} \gls{SSDL} training and thus improve the overall result. For reasons of space, the full results can be found in \Cref{tab:distances} of \Cref{app:exp-sim}.

As for the qualitative test results, Figure \ref{tab:Feature_plots-selected} shows an example of a feature density function approximated for randomly selected samples for the MNIST-ImageNet and  MNIST-\gls{SVHN} data set pairs. The plots reveal a stronger density based similarity between the MNIST and ImageNet than the MNIST and \gls{SVHN} data sets. This is in spite of perhaps higher semantic similarity between \gls{SVHN} and MNIST. This correlates well with the quantitative figures yielded in Table \ref{tab:distances}, where the MNIST data set is more dissimilar to the \gls{SVHN} data set (MNIST contaminated by 100\% with the \gls{SVHN} data set) than the ImageNet data set (MNIST contaminated by 100\% with the ImageNet data set). This also highly correlates with the final \gls{SSDL} accuracy yielded with both unlabelled data sets (MNIST contaminated by 100\% with \gls{SVHN}  and ImageNet) shown in Table \ref{tab:half-half-results}. MixMatch shows a higher accuracy when using ImageNet as an unlabelled data set compared to using \gls{SVHN}  as unlabelled data. For reasons of space, the full results can be found in \Cref{tab:Feature_plots}.
\section{Discussion}
In this work, we introduced the \textit{non-\gls{IID}-SSDL} sandbox. We demonstrated that semantic similarity matching between labelled and unlabelled data is not a reliable recipe for successful \gls{SSDL}. We also showed that simple dissimilarity measurements in the feature space of a generic classifier offer a promising and reproducible alternative.

When relaying these results back to application domains we consider the following implications. Real-world \gls{SSDL} may come with different degrees of \gls{OOD} data contamination: for instance, with a deep learning model trained for medical imaging analysis, unlabelled data can include images within the same domain, but capturing different pathologies not present in the labelled data. A close scenario has been simulated with the \gls{OH} setting which resulted in a subtle accuracy degradation in most cases. However, the accuracy gain obtained in our tests vis-a-vis the fully supervised baseline is still substantial.
Another plausible real-world scenario for \gls{SSDL} is the inclusion of widely available unlabelled data sets, e.g. built with web crawlers, where the domain shift can be more substantial. Such scenarios have been simulated with the \gls{OOD} types similar and different. We can observe that notions of semantic similarity between labelled and unlabelled data set pairings, e.g. (MNIST-SVHN) or (FashionMNIST-Fashion Product), do not necessarily imply an \gls{SSDL} accuracy gain. Dissimilarity measures, in particular $d_C$, show promise as a simple proxy for \gls{SSDL} accuracy, according to our test results. This is visible when comparing the accuracy and distance results of the previous pairings to (MNIST-Tiny ImageNet) and (FashionMNIST-Tiny ImageNet) which have higher accuracies and, also, surprisingly, lower distance measurements. We speculate how this can be related to the quality of the feature extractor that the usage of more diverse data can yield, as the positive results often yielded with self-supervised learning \citep{zhai2019s4l}. 

In future work, we plan to investigate the relationship between generic feature similarity and \gls{SSDL} downstream performance further. The fact that feature dissimilarity scores can be calculated before \gls{SSDL} training and independent of the \gls{SSDL} model poses an interesting profile for application in data augmentation \citep{hendrycks2020augmix} or the auditing of machine learning algorithms \citep{pmlr-v136-oala20a}. Finally, further exploration of the connections to \gls{OOD} detection \citep{zisselman2020deep}, concept drift \citep{webb2016characterizing} and class distribution mismatch \citep{chen2020semi} offer interesting avenues for further research in non-IID \gls{SSDL}.
\subsubsection*{Acknowledgments}
We would like to thank Jordina Torrents-Barrena and Sören Becker for fruitful discussions during the inception of this work. We would like to thank Shengxiang Yang, Armaghan Moemeni, Wojciech Samek and Miguel A. Molina-Cabello for feedback on the first manuscript.
\bibliographystyle{iclr2021_conference}
\bibliography{mendeley,luis_refs,sauls_references,related_work}
\newpage
\appendix
\section{Method - Details}
\subsection{MixMatch Algorithm and Deep Learning Model}
\label{app:algo}
In MixMatch, the consistency loss term minimizes the distance of the pseudo-labels and the model predictions over the unlabelled data set $X_{u}$. Pseudo-label $\widehat{\boldsymbol{y}}{}_{j}$ estimation is performed with the average model output of a transformed input $x_{j}$, with $K$ number of different transformations.  $K=2$ is advised in \citep{berthelot2019mixmatch}. The estimated pseudo-labels $\widehat{\boldsymbol{y}}$ might be too \textit{unconfident}. To tackle this, pseudo-label sharpening is performed with a temperature $\rho$.  The data set with the estimated and sharpened pseudo-labels was defined as $\widetilde{S}_{u}=\left(X_{u},\widetilde{Y}\right)$, with $\widetilde{Y}=\left\{ \widetilde{\boldsymbol{y}}_{1},\widetilde{\boldsymbol{y}}_{2},\ldots,\widetilde{\boldsymbol{y}}_{n_{u}}\right\}$.

Data augmentation is a key aspect in semi-supervised deep learning as found in \citep{berthelot2019mixmatch}. To further augment data using both labelled and unlabelled samples, they implemented the Mix Up algorithm developed in \citep{zhang2017mixup}. Linear interpolation of a mix labelled observations and unlabelled (with its corresponding pseudo-labels) observations.
\begin{equation}
\left(S'_{l},\widetilde{S}'_{u}\right)=\Psi_{\textrm{MixUp}}\left(S_{l},\widehat{S}_{u},\alpha\right)
\end{equation}

The Mix Up algorithm creates new observations from a linear interpolation of a mix of unlabelled (with its corresponding pseudo-labels) and labelled data. More specifically, it takes two labelled (or pseudo labelled) data pairs $\left(\boldsymbol{x}_{a},y_{a}\right)$ and $\left(\boldsymbol{x}_{b},y_{b}\right)$. The Mix Up method generates a new observation and its label $\left(\boldsymbol{x}',y'\right)$ by following these steps: 

\begin{enumerate}
\item  Sample the Mix Up parameter $\lambda$ from a Beta distribution $\lambda\sim\textrm{Beta}\left(\alpha,\alpha\right)$.

\item  Ensure that $\lambda>0.5$ by making $\lambda'=\max\left(\lambda,1-\lambda\right)$.

\item  Create a new observation with a lineal interpolation of both observations: $\boldsymbol{x}'=\lambda'\boldsymbol{x}_{a}+\left(1-\lambda'\right)\boldsymbol{x}_{b}$.
\end{enumerate}

With the augmented data sets $ \left(S'_{l},\widetilde{S}'_{u}\right)$, the MixMatch training can be summarized as:
\begin{equation}
f_{\boldsymbol{w}}=T_{\textrm{MixMatch}}\left(S_{l},S_{u},\alpha,\gamma, \lambda\right)=\underset{\boldsymbol{w}}{\textrm{argmin}}\mathcal{L}\left(S,\boldsymbol{w}\right)
\end{equation}

\begin{equation}
\mathcal{L}\left(S,\boldsymbol{w}\right)=\sum_{\left(\boldsymbol{x}_{i},\boldsymbol{y}_{i}\right)\in S'_{l}}\mathcal{L}_{l}\left(\boldsymbol{w},\boldsymbol{x}_{i},\boldsymbol{y}_{i}\right)+
r(t) \gamma\sum_{\left(\boldsymbol{x}_{j},\widetilde{\boldsymbol{y}}_{j}\right)\in\widetilde{S}'_{u}}\mathcal{L}_{u}\left(\boldsymbol{w},\boldsymbol{x}_{j},\widetilde{\boldsymbol{y}}_{j}\right)
\end{equation}

For  supervised and semi-supervised loss functions,  the cross-entropy $\mathcal{L}_{l}\left(\boldsymbol{w},\boldsymbol{x}_{i},\boldsymbol{y}_{i}\right)=\delta_{\textrm{cross-entropy}}\left(\boldsymbol{y}_{i},f_{\boldsymbol{w}}\left(\boldsymbol{x}_{i}\right)\right)$ and the Euclidean distance $\mathcal{L}_{u}\left(\boldsymbol{w},\boldsymbol{x}_{j},\widetilde{\boldsymbol{y}}_{j}\right)=\left\Vert \widetilde{\boldsymbol{y}}_{j}-f_{\boldsymbol{w}}\left(\boldsymbol{x}_{j}\right)\right\Vert$, are usually implemented, respectively. The regularization  $\gamma$ controls the direct influence on unlabelled data. Since in the first epochs, unlabelled data based predictions are unreliable, the function $r(t) = t/\rho$ increases the unsupervised term contribution as the number of epochs progress.  The coefficient $\rho$ is referred as the rampup coefficient.  Unlabelled data also influences the labelled data term $\mathcal{L}_{l}$, since unlabelled data is used also to artificially augment the data set with the Mix Up algorithm. This loss term is used at training time, for testing, a regular cross entropy loss is implemented.
\subsection{Dissimilarity Measures}
\label{app:measures}
In this work we implement a set of \gls{DeDiM}s based on data set subsampling, as image data sets are usually large, following a sampling approach for comparing two populations, as seen in \citep{krzanowski2003non}. We compute the dissimilarity measures in the feature space of a generic Wide-ResNet pre-trained on ImageNet, making our proposed approach non-specific to the \gls{SSDL} model to be trained. This enables an evaluation of the unlabelled data before training the \gls{SSDL} model. The proposed measures in this work are meant to be simple and quick to evaluate with practical use in mind. We propose and test the implementation of two Minkowski based distance sets, $d_{\ell_{2}}\left(S_{a},S_{b},\tau, \mathcal{C}\right)$ and $d_{\ell_{1}}\left(S_{a},S_{b},\tau, \mathcal{C}\right)$, corresponding to the Euclidean and Manhattan distances, respectively, between two data sets $S_a$ and $S_b$. Additionally, we implement and test two non-parametric density based data set divergence measures; Jensen-Shannon ($d_{\textrm{JS}}$) and cosine distance ($d_{C}$). 
For all the proposed dissimilarity measures, the parameter $\tau$ defines the sub-sample size used to compute the dissimilarity between the two data sets $S_a$ and $S_b$, and $\mathcal{C}$ the total number of samples to compute the mean sampled dissimilarity measure. The general procedure for all the implemented distances is detailed as follows. 

\begin{itemize}[leftmargin=*]
\item We randomly sub-sample each one of the  data sets $S_a$ and $S_b$, with a sample size of $\tau$, creating the sampled data sets $S_{a,\tau}$ and $S_{b,\tau}$.

\item We transform an input observation $\boldsymbol{x}_j\in S_i$, with $\boldsymbol{x}_j \in \mathbb{R}^{n}$, being $n$  the dimensionality of the input space, using the feature extractor $f$, yielding the feature vector $ \boldsymbol{h}_{j}=f\left(\boldsymbol{x}_{j}\right)$. 

\item The feature vector $\boldsymbol{h}_{i}  \in \mathbb{R}^{n'}$ has dimension $n'$  dimensions, with $n' < n$. For instance, the implemented feature extractor $f$ uses the ImageNet pretrained  Wide-ResNet architecture, extracting $n'=512$ features. This yields the two feature sets $H_{a,\tau}$ and $H_{b,\tau}$

\end{itemize}

For the Minkowski based distance sets $d_{\ell_{2}}\left(S_{a},S_{b},\tau, \mathcal{C}\right)$, $d_{\ell_{1}}\left(S_{a},S_{b},\tau, \mathcal{C}\right)$, we perform the following steps for the sets of features obtained in the previous description $H_{a,\tau}$ and $H_{b,\tau}$:

\begin{itemize}[leftmargin=*]
    \item For each feature vector $\boldsymbol{h}_j \in H_{a,\tau}$, find the closest feature vector $\boldsymbol{h}_k \in H_{b,\tau}$, using the $\ell_{p}$  distance, with $p=1$ or $p=2$ for the Manhattan and Euclidean distances, respectively:  
    $\widehat{d}_{j}={\textrm{min}}_k\left\Vert \boldsymbol{h}_{j}-\boldsymbol{h}_{k}\right\Vert _{p}.$ We do this for a number of $\mathcal{C}$ samples, yielding a list of distance calculations $d_{\ell_{p}}\left(S_{a},S_{b},\tau, \mathcal{C}\right) = \left\{ \widehat{d}_{1}, \widehat{d}_{2},..., \widehat{d}_\mathcal{C} \right\} $. 
    
    \item We compute a reference list of distances for the same list of samples of the data set $S_a$ to itself (intra-data set distance), thereby computing $d_{\ell_{p}}\left(S_{a},S_{a},\tau, \mathcal{C}\right)$. This yields a list of reference distances $\check{d}_{1}, \check{d}_{2}, ..., \check{d}_{\mathcal{C}}$. In our case $S_a$ corresponds to the labelled data set $S_l$, as the distance to different unlabelled data sets $S_u$ is to be computed.
    
    \item To ensure that the absolute differences between the reference and inter-data set distances $d_{c}=\left|\widehat{d}_{c}-\check{d}_{c}\right|$ are statistically significant, we compute the $p$-value associated with a Wilcoxon test. 
    
    \item After the distance set between two data sets $d_{\ell_{p}}\left(S_{a},S_{b},\tau, \mathcal{C}\right)$ is obtained, its average reference subtracted distance $\overline{d}$ and its corresponding statistical significance $p$-value are computed. 
\end{itemize}
As for the density based distances implemented we follow a similar sub-sampling approach, with these steps:
\begin{itemize}[leftmargin=*]
    \item For each dimension $r=1,...,n'$ in the feature space, we compute the normalized histograms to approximate the density functions $p_{r,a}$, in the sample $H_{a,\tau}$. Similarly, we compute the normalized histograms to yield the set of approximate density functions $p_{r,b}$ for $r=1,...,n'$, using the observations in the sample $H_{b,\tau}$. 
    \item For the Jensen-Shannon divergence ($d_{\textrm{JS}}$) and the cosine distance ($d_{C}$), we compute the sum of the dissimilarities between the density functions $p_{r,a}$ and $p_{r,b}$, to yield the estimated dissimilarity for the sample $j$: 
    $\widehat{d}_{j}=\ensuremath{\sum_{r=1}^{n'}\delta_{g}\left(p_{r,a},p_{r,b}\right)}$, where $g=JS$ and $g=C$ for the Jensen-Shannon divergence and the cosine distance, respectively. 
    We do this for all the $\mathcal{C}$ samples, yielding the list of inter-data set distances: $\widehat{d}_{1}, \widehat{d}_{2},..., \widehat{d}_\mathcal{C}$. To lower the computational burden,  we assume that the dimensions are statistically independent. 
    
    \item Similar to the Minkowski distances, we compute the intra-data set distances for the data set $S_a$, in this context the labelled data set $S_l$, to obtain the list of reference distances $\check{d}_{1}, \check{d}_{2}, ..., \check{d}_{\mathcal{C}}$.
    
    \item Similarly, to verify that the inter- and intra-data set distance differences  $d_{c}=\left|\widehat{d}_{c}-\check{d}_{c}\right|$ are statistically significant, we compute the $p$-value associated with a Wilcoxon test. The distance computation yields the sample mean distance   $\overline{d}$ and its statistical significance $p$-value.
\end{itemize}

The proposed dissimilarity measures do not fulfill the conditions to be a mathematical pseudo-metric since the distance of an object to itself is not exactly zero  and symmetry properties are not fulfilled for the sake of evaluation speed \citep{cremers2003pseudo}. Despite these relaxations, we will see that these dissimilarity measures, especially the two that are density based, are an effective proxy for estimating the  $S_{\textrm{u,OOD}}$ accuracy gain. 
\section{Experiments - Details}
\subsection{Hyperparameters}
\label{app:hyper}
\subsubsection{Global hyperparameters}
\begin{table}[h]
\caption{Global hyperparameters which are kept constant throughout all experiments. This was done in order to isolate the effects of the changing \gls{OOD} data configurations.}
\centering
\begin{tabular}{cc}
\toprule
\textbf{Description}             & \textbf{Value} \\
\midrule
Model architecture used in all tasks                & wide\_resnet           \\
Number of training epochs            & $50$          \\
Batch size & $16$     \\ 
Learning rate & $0.0002$\\
Weight decay & $0.0001$\\
Rampup coefficient & $3000$\\
Optimizer & Adam with\\
& 1-cycle policy \citep{smith2018disciplined}\\
$K$, Number of augmentations                & $2$            \\
$T$, Sharpening temperature            & $0.5$          \\
$\alpha$, Parameter for the Beta distribution & $0.75$     \\ 
$\gamma$, Gamma for the unsupervised loss weight  & $25$ \\
\bottomrule
\end{tabular}
\vspace{1mm}
\label{table:globalhyper}
\end{table}
\subsubsection{MixMatch hyperparameters}
For  supervised and semi-supervised loss functions, the cross-entropy and the Euclidean distance, are used, respectively. The regularization coefficient  $\gamma$ controls the direct influence on unlabelled data. Unlabelled data also influences the labelled data term since unlabelled data is used also to artificially augment the data set with the Mix Up algorithm. This loss term is used at training time, for testing, a regular cross entropy loss is implemented. We use the recommended hyperparameters documented in \citep{berthelot2019mixmatch}. 
\begin{table}[h]
\caption{MixMatch hyperparameters. All parameters were chosen following the recommendations by \citep{berthelot2019mixmatch}.}
\centering
\begin{tabular}{cccc}
\toprule
\textbf{Symbol} & \textbf{Description} & \textbf{Name in code}             & \textbf{Value} \\
\midrule
$K$                & Number of augmentations                & \url{K_TRANSFORMS}& $2$            \\
$T$                & Sharpening temperature            & \url{T_SHARPENING} & $0.5$          \\
$\alpha$           & Parameter for the Beta distribution & \url{ALPHA_MIX} & $0.75$     \\ 
$\gamma$ & Gamma for the loss weight & \url{GAMMA_US} & $25$ \\
- & Whether to use balanced (5) & \url{BALANCED} & $5$\\
 &  or unbalanced (-1) loss for MixMatch & & \\
\bottomrule
\end{tabular}
\vspace{1mm}
\label{table:mixmatchhyper}
\end{table}
\subsection{Data sets overview}
\label{app:data sets}
If you wish to reproduce any of the experiments data sets are automatically downloaded by the experiment script \url{ood_experiment_at_scale_script.sh} for your convenience based on which experiment you choose to run.

An overview of the different data sets can be found below. Note that we used the training split of each data set as the basis to construct our own training and test splits for each experimental run.

\begin{table}[h]
\caption{Information on the data sets used in the experiments. \textbf{Format} specifies the format the image files were provided in, $\mathbf{d}$ specifies the size of the images, $\mathbf{N}$ specifies the number of samples in the data set, $\mathbf{|\mathcal{Y}|}$ specifies the number of classes in the data set, \textbf{Relative class distribution} specifies the relative class distribution in the data set.}
\centering
\scalebox{0.8}{
    \begin{tabular}{cccccc}
    \toprule
    \textbf{data set} & \textbf{Format} & $\mathbf{d}$  & $\mathbf{N}$ & $\mathbf{|\mathcal{Y}|}$ & \textbf{Relative class distribution} \\
    \midrule
    \textbf{MNIST} \citep{lecun1998gradient} & \url{.jpg} & $28 \times 28$ & 42,000 & 10 & Uniform\\
    \multirow{2}{*}{\textbf{SVHN} \citep{netzer2011reading}} &\url{.png} & $32 \times 32$ & 73,557 & 10 & 0.07/0.19/0.14/0.12/0.1/ \\
    & & & & & 0.09/0.08/0.07/0.07/0.07 \\
    \textbf{Tiny ImageNet} \citep{deng2009imagenet} & \url{.jpg} & $64 \times 64$ & 100,000 & 200 & Uniform\\
    \textbf{CIFAR-10} \citep{krizhevsky2009learning} & \url{.jpg} & $32 \times 32$ & 50,000 & 10 & Uniform\\
    \textbf{FashionMNIST} \citep{xiao2017} & \url{.png} & $28 \times 28$& 60,000 & 10 & Uniform\\
    \textbf{Fashion Product} \citep{fashionproduct} & \url{.jpg} & $60 \times 80$ & 44,441 & 5 & 0.48/0.25/0.21/0.05/0.01 \\
    \textbf{Gaussian} & \url{.png} & $224 \times 224$& 20,000 & NA & NA\\
    \textbf{Salt and Pepper} & \url{.png} & $224 \times 224$& 20,000 & NA & NA\\
    \bottomrule
    \end{tabular}
}
\vspace{1mm}
\label{table:data set}
\end{table}
The Gaussian and Salt and Pepper data sets were created with the following parameters: a variance of 10 and mean 0 for the Gaussian noise, and an equal Bernoulli probability for 0 and 255 pixels, in the case of the Salt and Pepper noise. 
\subsection{Preprocessing}
Each data point was preprocessed in the following way. After a subset of labelled and unlabelled data for an experimental run had been constructed the means and standard deviations (one pair for labelled data, one pair for unlabelled data) were calculated for this specific subset. Then, the labelled and unlabelled inputs were standardized by subtracting the respective mean and dividing by the respective standard deviation.

In addition, in situations when the size of the unlabelled images differed from the size of the labelled images up- or downsampling was used to align the unlabelled image size.
\subsection{Hardware}
\label{app:hardware}
Experiments were run on three machines. Machine 1 has one 12GB NVIDIA TITAN X GPU, 24 Intel(R) Xeon(R) CPU E5-2687W v4 @ 3.00GHz and 32GB RAM. Machine 2 has four 16GB NVIDIA T4 GPUs, 44 CPUs from the Intel Xeon Skylake family and 150GB RAM. Machine 3 has one 12GB NVIDIA TITAN V GPU, 24 Intel(R) Xeon(R) E5-2620 0 @ 2.00GHz CPU and 32GB RAM. Experimental runs were parallelized using the ampersand option in \url{bash} executing 10 runs in parallel on a single GPU. With the current code base this requires up to 10 CPUs per GPU as well as approximately 25GB RAM per GPU. With this setup a single training epoch of 10 parallel experimental runs should last between 2 and 4 minutes per GPU, depending on which type of GPU is used. 
\newpage
\subsection{\textit{non-\gls{IID}-\gls{SSDL}} Sandbox - Full Results}
\label{app:exp-sandbox}
\begin{table*}[h]
\caption{Results for the class distribution mismatch experiment, best \gls{OOD} performance in bold per configuration. Each result entry in the table represents the mean and variance of accuracy across ten random experimental runs per entry. For a detailed description of symbols and the experiment see \Cref{sec:experiments}. Numbers in parentheses next to $S_\textrm{u,OOD}$ entries reflect the preference ranking provided by the dissimilarity measure with $d_{C}$.}
\label{tab:half-half-results}

\centering

\scalebox{0.8}{

\begin{tabular}{ccccccc}
\toprule
\multirow{2}{*}{$\mathbf{S_{IOD}}$} & \multirow{2}{*}{$\mathbf{T_{OOD}}$} & \multirow{2}{*}{$\mathbf{S_{uOOD}}$} & \multirow{2}{*}{$\mathbf{\%_{uOOD}}$} & \multicolumn{3}{c}{$\mathbf{n_l}$}\\

&  & &  & 60 & 100 & 150\\

\midrule

\multirow{12}{*}{\rotatebox[origin=c]{90}{\textbf{MNIST}}} 
& \multicolumn{3}{c}{Fully supervised baseline} & $0.457\pm0.108$ & $0.559\pm 0.125 $ & $0.645 \pm0.101$ \\

 & \multicolumn{3}{c}{SSDL baseline (no OOD data)} & $0.704 \pm 0.096$ & $0.781\pm 0.065$ & $0.831 \pm0.0626$\\

 \cline{2-7}
 
 & \multirow{2}{*}{OH} & \multirow{2}{*}{MNIST} & 50 & $\mathbf{0.679\pm0.108}$ & $\mathbf{0.769\pm0.066}$ & $0.802\pm0.054$\\
 & & & 100 & $0.642 \pm 0.111$ & $0.746 \pm0.094$ & $0.798 \pm0.07$\\

 \cline{2-7}
 & \multirow{2}{*}{Sim} & \multirow{2}{*}{SVHN (3)} & 50 & $0.637 \pm0.098$ & $0.745 \pm0.081$ & $0.801 \pm0.0699$\\
 
 & & & 100 & $0.482 \pm0.113$ & $0.719 \pm0.058$ & $0.765 \pm0.072$\\
 
  \cline{2-7}
 
  & \multirow{6}{*}{Dif}& \multirow{2}{*}{TI (2)} & 50 & $0.642 \pm0.094$ & $0.739 \pm 0.074$ & $0.809 \pm0.066$\\
  
  & & & 100 & $0.637 \pm0.097$ & $0.732\pm 0.074$ & $0.804 \pm0.071$\\

 \cline{3-7}
 
  & & \multirow{2}{*}{GN (5)} & 50 & $0.606 \pm0.0989$ & $0.713 \pm0.087$ & $0.786\pm 0.065$\\
  
  & & & 100 & $0.442 \pm0.099$ & $0.461 \pm0.073$ & $0.542\pm 0.062$\\

   \cline{3-7}
 
  & & \multirow{2}{*}{SAPN (4)} & 50 & $0.631 \pm0.102$ & $0.735 \pm0.082$ & $\mathbf{0.813 \pm0.057}$\\
  
  & & & 100 & $0.48\pm0.0951$ & $0.524 \pm0.09$ & $0.563 \pm0.095$\\
 
\hline

\multirow{12}{*}{\rotatebox[origin=c]{90}{\textbf{CIFAR-10}}} & \multicolumn{3}{c}{Fully supervised baseline} & $0.380\pm0.024$ & $0.445\pm0.042$ & $0.449\pm0.022$\\

 &  \multicolumn{3}{c}{SSDL baseline (no OOD data)} & $0.453\pm0.046$ & $0.474\pm0.019$ & $0.501\pm0.033$\\

 \cline{2-7}

& \multirow{2}{*}{OH} & \multirow{2}{*}{CIFAR-10} & 50 & $\mathbf{0.444\pm0.040}$ & $0.472\pm0.039$ & $0.525\pm0.050$\\ 
 & & & 100 & $0.443\pm0.023$ & $0.472\pm0.047$ & $0.499\pm0.054$ \\
 \cline{2-7}
 & \multirow{2}{*}{Sim} & \multirow{2}{*}{TI (1)} & 50 & $0.435\pm0.054$ & $0.473\pm0.039$ & $\mathbf{0.543\pm0.040}$\\ 
 & & & 100 & $0.417\pm0.020$ & $\mathbf{0.480\pm0.039}$ & $0.498\pm0.042$ \\
 
 \cline{2-7}
 
  & \multirow{6}{*}{Dif} & \multirow{2}{*}{SVHN (2)} & 50 & $0.419\pm0.027$ & $0.464\pm0.044$ & $0.469\pm0.056$\\ 
 & & & 100 & $0.385\pm0.034$ & $0.418\pm0.035$ & $0.440\pm0.046$ \\
 
 \cline{3-7}
 
  & & \multirow{2}{*}{GN (4)} & 50 & $0.409\pm0.047$ & $0.454\pm0.048$ & $0.491\pm0.032$\\ 
 & & & 100 & $0.297\pm0.029$ & $0.306\pm0.034$ & $0.302\pm0.038$ \\
 
 \cline{3-7}
 
  & & \multirow{2}{*}{SAPN (5)}& 50 & $0.438\pm0.029$ & $0.455\pm0.037$ & $0.485\pm0.034$\\ 
 & & & 100 & $0.236\pm0.031$ & $0.246\pm0.032$ & $0.232\pm0.022$ \\
  
  \hline

\multirow{12}{*}{\rotatebox[origin=c]{90}{\textbf{FashionMNIST}}} & \multicolumn{3}{c}{Fully supervised baseline} & $0.571\pm0.073$ & $0.704\pm0.066$ & $0.720\pm0.093$\\

 & \multicolumn{3}{c}{SSDL baseline (no OOD data)} & $0.715\pm0.049$ & $0.760\pm0.044$ & $0.756\pm0.069$\\
 \cline{2-7}

& \multirow{2}{*}{OH} & \multirow{2}{*}{FashionMNIST} & 50 & $\mathbf{0.714\pm0.049}$ & $0.721\pm0.104$ & $0.765\pm0.053$\\ 
 & & & 100 & $0.660\pm0.061$ & $0.711\pm0.090$ & $0.747\pm0.061$ \\
 \cline{2-7}
 & \multirow{2}{*}{Sim} & \multirow{2}{*}{FP (4)} & 50 & $0.707\pm0.039$ & $0.724\pm0.030$ & $0.778\pm0.078$\\ 
 & & & 100 & $0.546\pm0.101$ & $0.542\pm0.099$ & $0.540\pm0.105$ \\
 
 \cline{2-7}
 
  & \multirow{6}{*}{Dif}& \multirow{2}{*}{TI (2)} & 50 & $0.690\pm0.065$ & $\mathbf{0.745\pm0.093}$ & $0.792\pm0.058$\\ 
 & & & 100 & $0.690\pm0.073$ & $0.728\pm0.066$ & $\mathbf{0.794\pm0.056}$ \\
  \cline{3-7}
  & & \multirow{2}{*}{GN (5)} & 50 & $0.644\pm0.061$ & $0.689\pm0.075$ & $0.755\pm0.055$\\ 
 & & & 100 & $0.352\pm0.025$ & $0.366\pm0.065$ & $0.361\pm0.057$ \\
  \cline{3-7}
  & & \multirow{2}{*}{SAPN (3)} & 50 & $0.671\pm0.072$ & $0.708\pm0.095$ & $0.729\pm0.088$\\ 
 & & & 100 & $0.276\pm0.069$ & $0.297\pm0.046$ & $0.283\pm0.059$\\
 \bottomrule
\end{tabular}

}
\end{table*}
\newpage
\subsection{dissimilarity measures - Full Results}
\label{app:exp-sim}
\begin{table*}[h]
\caption{Distance measures between the labelled and unlabelled datasets $S_l$ and $S_u$. Numbers in italics correspond to results with $p>0.05$ for the Wilcoxon test. For a detailed description of symbols and the experiment see \Cref{sec:experiments}}
\label{tab:distances}
\begin{center}
\scalebox{0.8}{
\begin{tabular}{ccccccc} 
\toprule
$\mathbf{S_l}$     & $\mathbf{S_u}$  & $\mathbf{\%_{uOOD}}$  & $\mathbf{d_{\ell_2}}$  & $\mathbf{d_{\ell_1}}$  & $\mathbf{d_{JS}}$                      & $\mathbf{d_{C}}$                        \\ 
\midrule
\multirow{10}{*}{\rotatebox[origin=c]{90}{\textbf{MNIST}}}           & \multirow{2}{*}{OH}          & 50                   & $\mathit{0.011 \pm0.006}$         & $\mathit{0.459 \pm0.28}$          & $\mathit{0.266 \pm0.221}$                        & $\mathit{0.811 \pm0.512}$                         \\    
& & 100                   & $\mathit{0.014\pm0.019}$        & $\mathit{0.38\pm0.507}$         & $1.001 \pm0.725$                       & $1.263\pm0.665$                         \\ 
\cline{2-7}
                            & \multirow{2}{*}{SVHN}            & 50                    & $\mathit{0.09\pm0.017}$         & $1.569\pm0.504$        & $6.789 \pm0.924$                       & $12.021\pm1.757$                        \\
                            &                 & 100                   & $0.25\pm0.053$         & $4.702 \pm 1.04$       & $52.349 \pm 3.292$                     & $42.026\pm4.31$                         \\ 
\cline{2-7}
                            & \multirow{2}{*}{TI}              & 50                    & $0.008\pm0.023$       & $1.519\pm0.223$        & $3.663 \pm0.772$                       & $5.512 \pm0.767$                        \\
                            &                 & 100                   & $0.217\pm 0.04$        & $4.3 \pm 0.636$        & $10.305 \pm1.667$                      & $15.18\pm2.698$                         \\ 
\cline{2-7}
                            & \multirow{2}{*}{GN}               & 50                    & $0.11 \pm 0.0219$      & $1.958\pm0.534$        & $14.785 \pm1.052$                      & $23.593\pm1.859$                        \\
                            &                 & 100                   & $0.357\pm0.081$        & $5.987 \pm 1.091$      & $52.349\pm4.253$                       & ~$86.21\pm3.471$~                       \\ 
\cline{2-7}
                            & \multirow{2}{*}{SAPN }             & 50                    & $0.089\pm0.0311$       & $2.479\pm0.7433$       & $15.116 \pm1.475$                      & $20.151 \pm1.619$                        \\
                            &                 & 100                   & $0.323 \pm 0.07$       & $6.308 \pm 1.366$      & ~$53.397 \pm4.253$                     & $77.456\pm4.474$                       \\ 
\midrule
\multirow{10}{*}{\rotatebox[origin=c]{90}{\textbf{CIFAR-10}}}           & \multirow{2}{*}{OH}       & 50                   & $\mathit{0.056 \pm0.023}$        & $\mathit{0.915 \pm0.934} $         & $\mathit{0.338 \pm0.325}$                       & $0.892 \pm0.402$                         \\         
& & 100                   &   $\mathit{0.061 \pm 0.04}$                     &    $0.769 \pm0.461$                    &$\mathit{0.451 \pm0.41 }$                                        &   $0.648 \pm0.407$                                      \\ 
\cline{2-7}
                            & \multirow{2}{*}{TI}              & 50                    & $0.082 \pm0.037$       & $\mathit{0.928 \pm0.815}$       & $\mathit{0.388 \pm0.243}$                       & $\mathit{0.423 \pm0.362}$                        \\
                            &                 & 100                   & $\mathit{0.056 \pm0.048}$      & $\mathit{0.992 \pm 0.517}$      & $\mathit{0.469 \pm0.426}$                       & $0.415 \pm0.232$                        \\ 
\cline{2-7}
                            & \multirow{2}{*}{SVHN}            & 50                    & $\mathit{0.055 \pm0.032}$       & $\mathit{0.948 \pm0.699}$       & $\mathit{0.665 \pm0.565}$                       & $\mathit{0.414\pm 0.357}$                        \\
                            &                 & 100                   & $\mathit{0.075 \pm 0.036}$      & $1.291 \pm0.925$       & $0.736 \pm 0.658$                      & $0.581 \pm0.343$                        \\ 
\cline{2-7}
                            & \multirow{2}{*}{GN}               & 50                    & $\mathit{0.107 \pm0.083}$       & $1.344 \pm 1.156$      & $1.708 \pm0.421$                       & $3.001 \pm0.696$                        \\
                            &                 & 100                   & $0.127 \pm0.087$       & $1.531 \pm0.767$       & $5.855 \pm0.552$                       & $8.703 \pm 0.926$                       \\ 
\cline{2-7}
                            & \multirow{2}{*}{SAPN }             & 50                    & $0.1146 \pm0.044$      & $1.854 \pm 0.894$      & $2.299\pm0.691$                        & $2.56 \pm0.762$                         \\
                            &                 & 100                   & $0.208 \pm0.05$      & $5.502 \pm 1.156$      & $8.225 \pm0.866$                       & $9.554 \pm0.489$                        \\ 
\midrule
\multirow{10}{*}{\rotatebox[origin=c]{90}{\textbf{FashionMNIST} }} & \multirow{2}{*}{OH}            & 50                   & $\mathit{0.02 \pm0.012}$        & $\mathit{0.34 \pm0.162 }$         & $\mathit{0.669 \pm0.566}$                       & $\mathit{0.575 \pm0.423}$                         \\    

& & 100                   &   $ 0.059 \pm0.032$ &  $0.801 \pm0.402$ & $0.305 \pm0.237$                   & $0.774 \pm0.343$                  \\ 
\cline{2-7}
                            & \multirow{2}{*}{FP}              & 50                    & $0.105 \pm 0.0526$     & $2.168\pm 0.774$       & $7.263 \pm 0.622$ & $5.305 \pm0.405$   \\
                            &                 & 100                   & $0.195\pm 0.0457$      & $4.819 \pm1.077$       & $9.056 \pm 0.462$ & $11.286 \pm0.751$  \\ 
\cline{2-7}
                            & \multirow{2}{*}{TI}              & 50                    & $\mathit{0.04 \pm0.03}$       & $\mathit{0.798 \pm 0.542}$      & $\mathit{0.897 \pm 0.516}$ & $0.897 \pm0.516$   \\
                            &                 & 100                   & $0.065 \pm0.03$      & $1.66 \pm 0.45$        & $1.4 \pm 0.488$   & $1.912 \pm0.683$    \\ 
\cline{2-7}
                            & \multirow{2}{*}{GN}               & 50                    & $\mathit{0.047 \pm 0.03}$       & $\mathit{0.533\pm 0.347}$       & $2.819 \pm 0.703$ & $3.843 \pm0.704$  \\
                            &                 & 100                   & $0.074 \pm0.041$       & $1.325 \pm0.631$       & $9.042 \pm 0.699$  & $15.511 \pm0.445$   \\ 
\cline{2-7}
                            & \multirow{2}{*}{SAPN }             & 50                    & $0.036 \pm0.022$       & $0.52 \pm0.303$        & $2.799 \pm 0.497$ & $2.799 \pm0.497$  \\
                            &                 & 100                   & $0.076 \pm0.044$       & $1.411 \pm0.548$       & $8.464 \pm 0.553$  & $8.464 \pm0.553$ \\
\bottomrule
\end{tabular}
}

\end{center}
\end{table*}

\newpage
\subsection{Feature Comparison - Additional Samples}
\label{app:feature}
\begin{table}[h]
        \centering
        \scalebox{0.91}{
            \begin{tabular}{cc}
               \centering
                  \scalebox{0.28}{\includegraphics{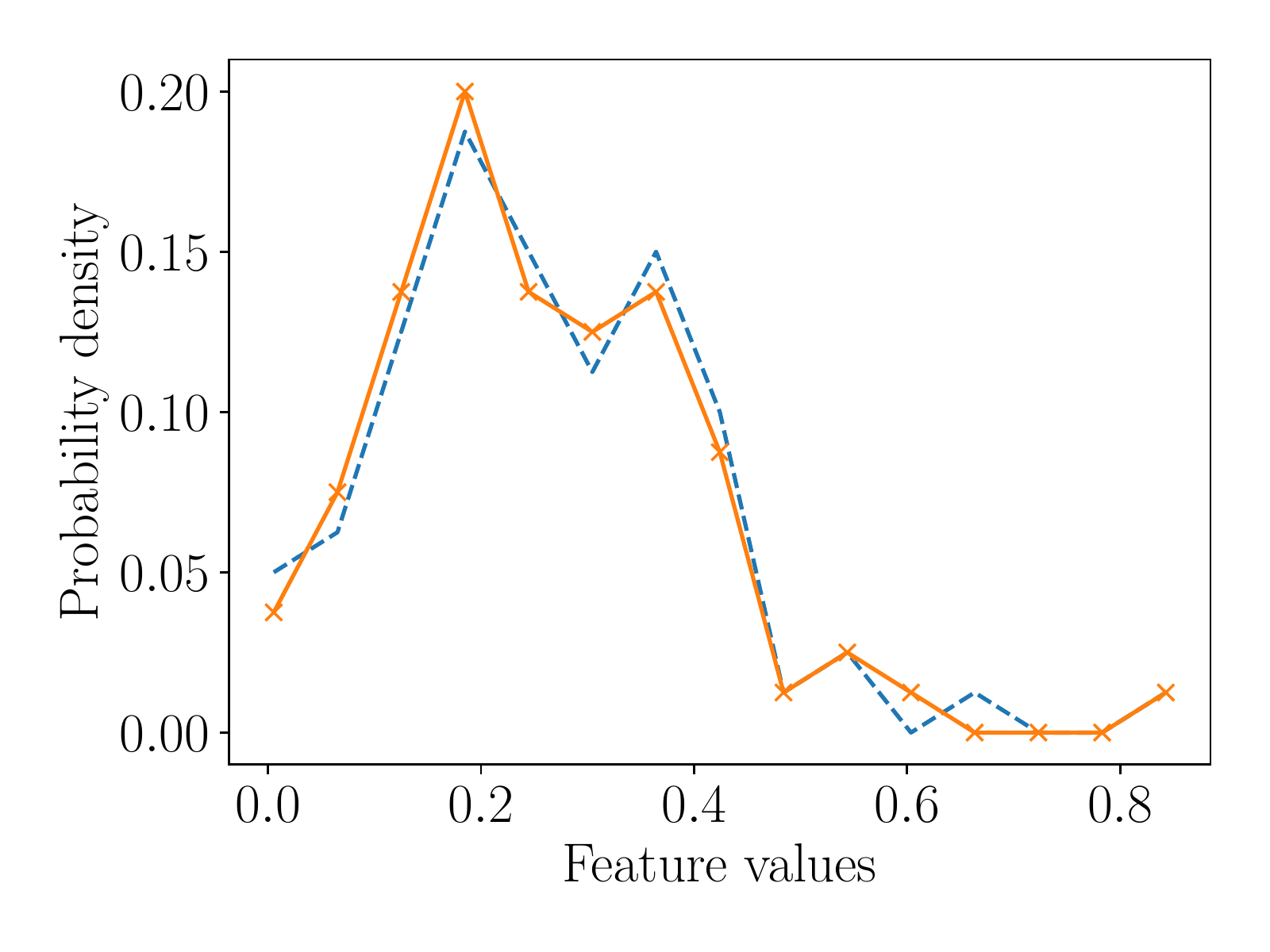}} & \scalebox{0.28}{\includegraphics{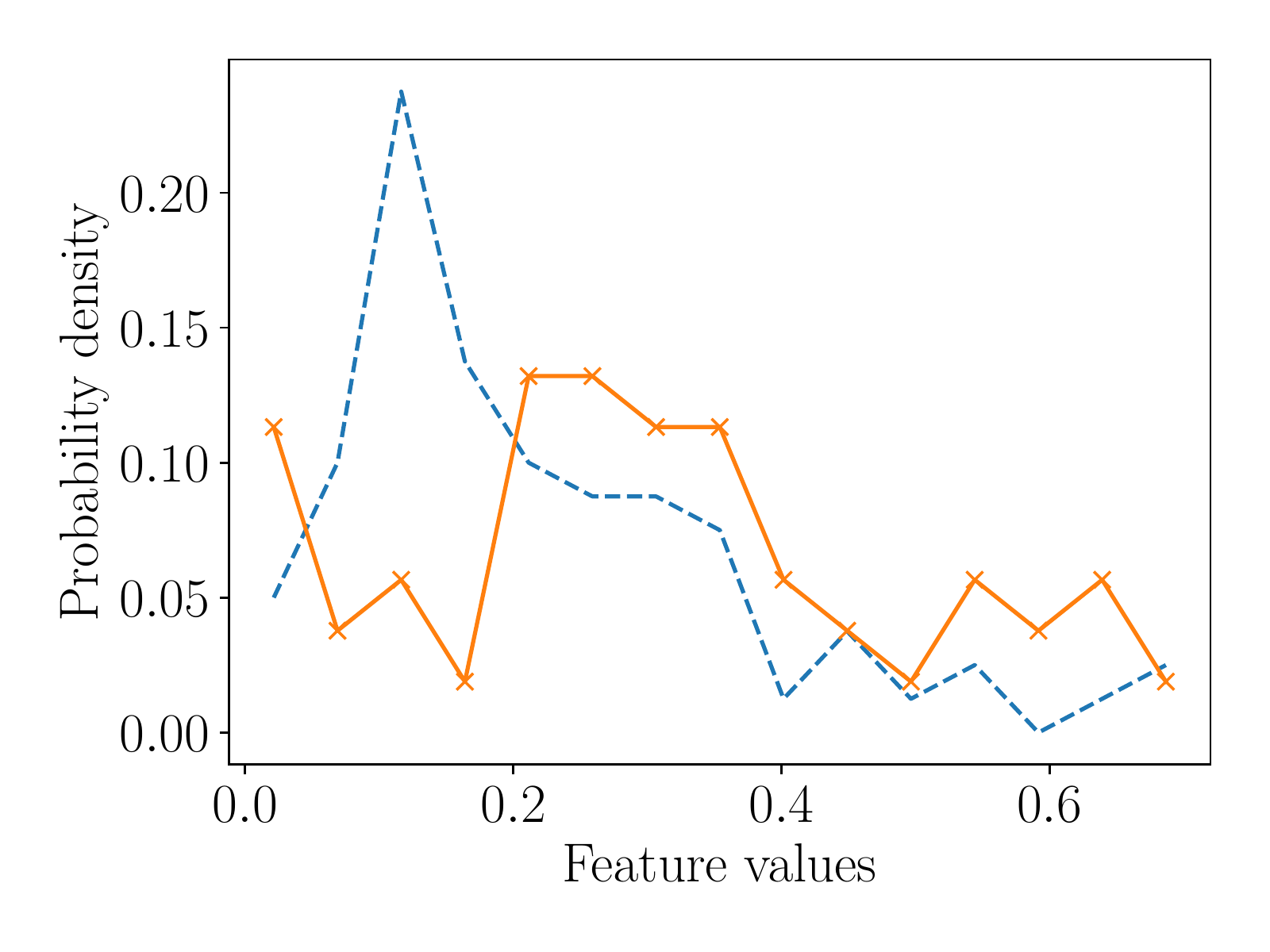}}  \\
                \scalebox{0.28}{\includegraphics{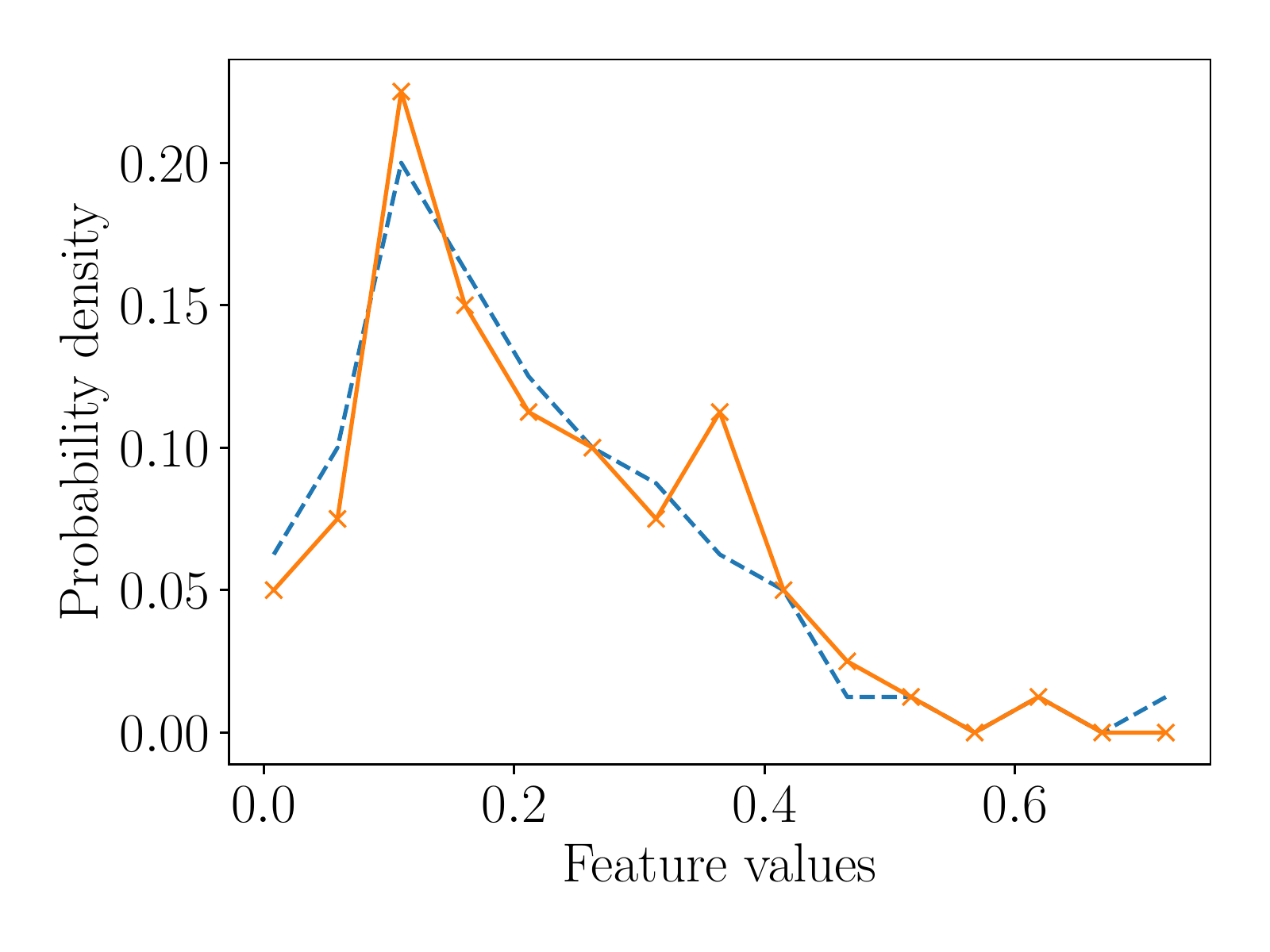}} & \scalebox{0.28}{\includegraphics{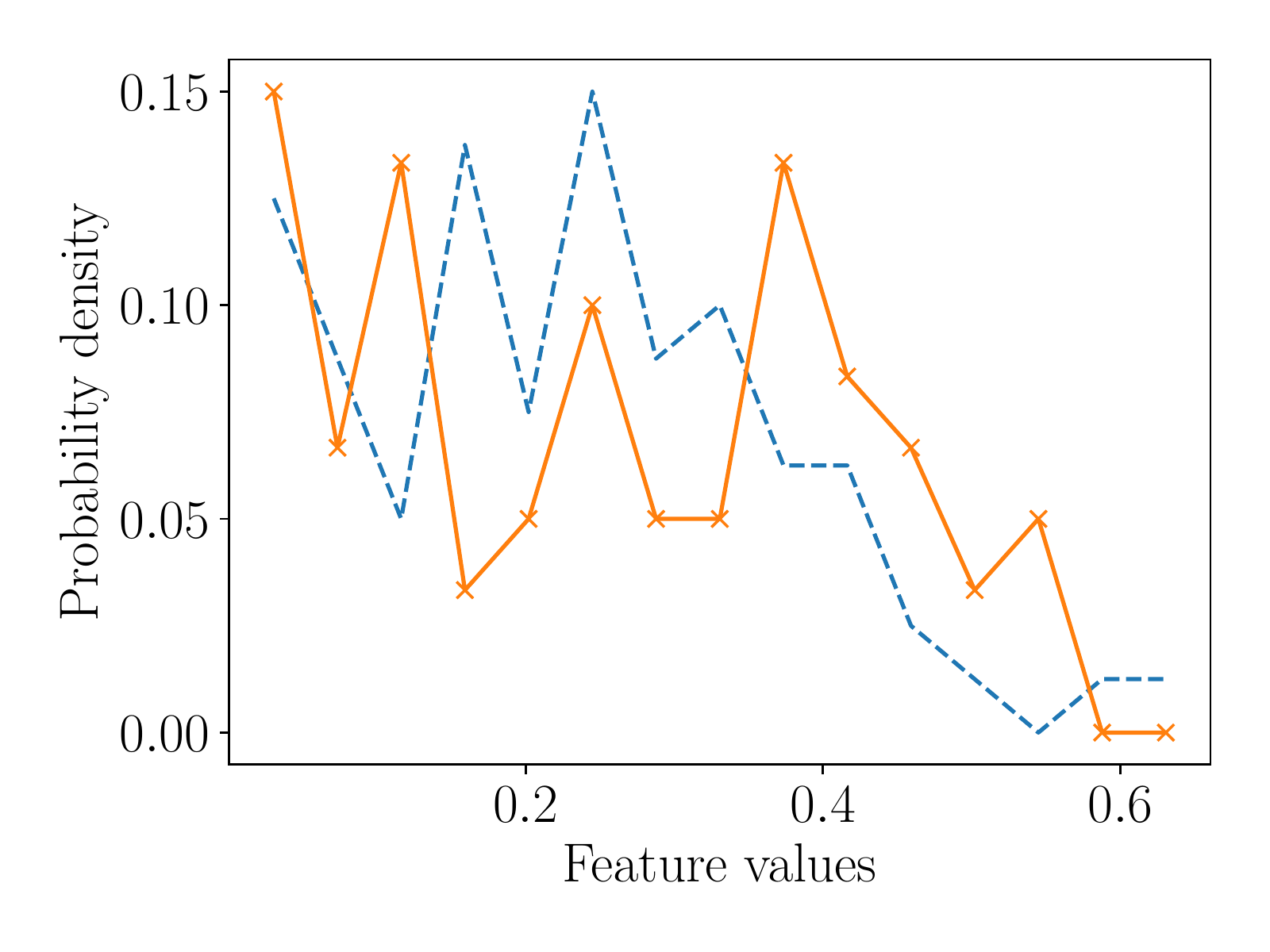}}\\
                
                \scalebox{0.28}{\includegraphics{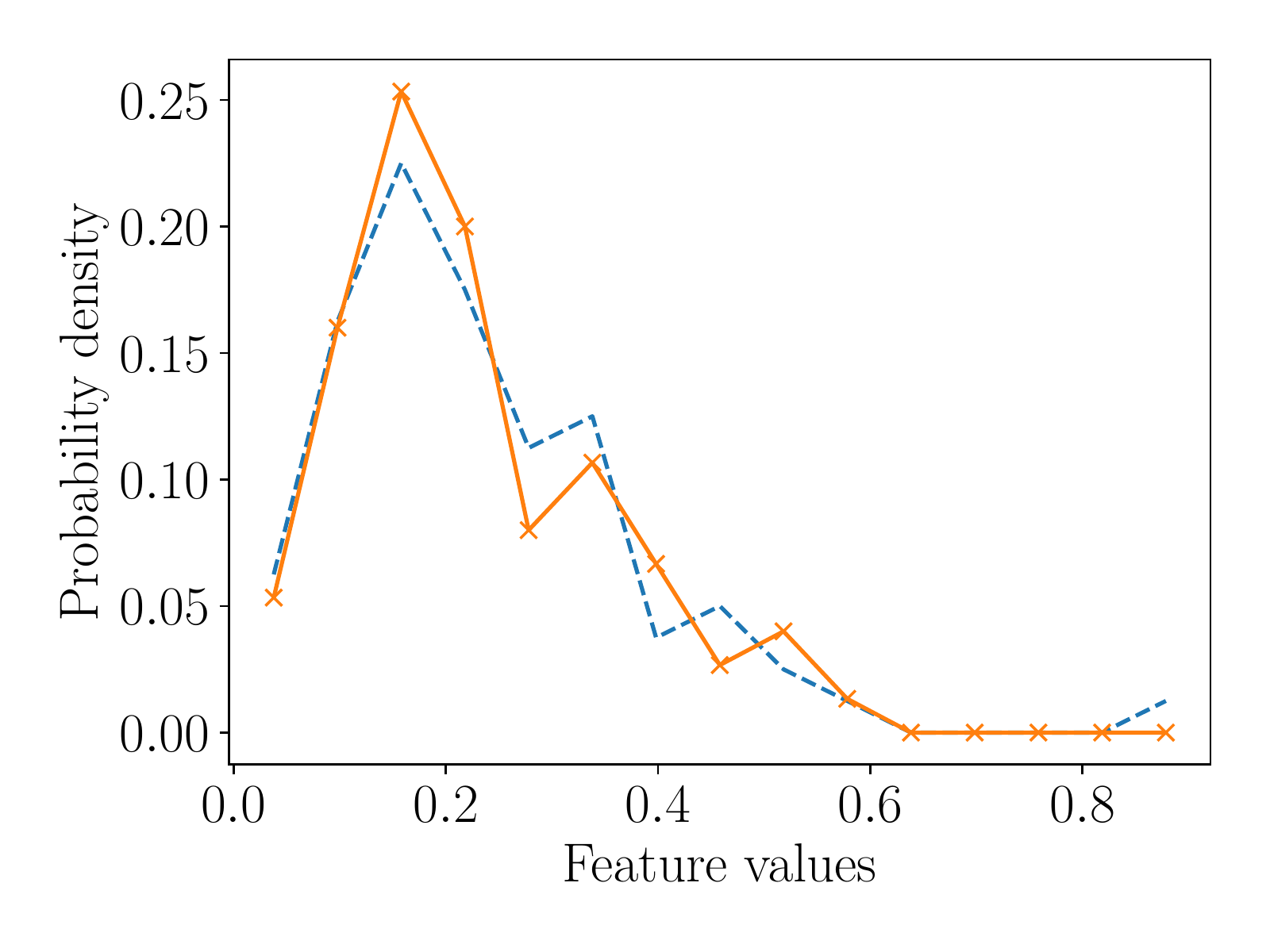}} & \scalebox{0.28}{\includegraphics{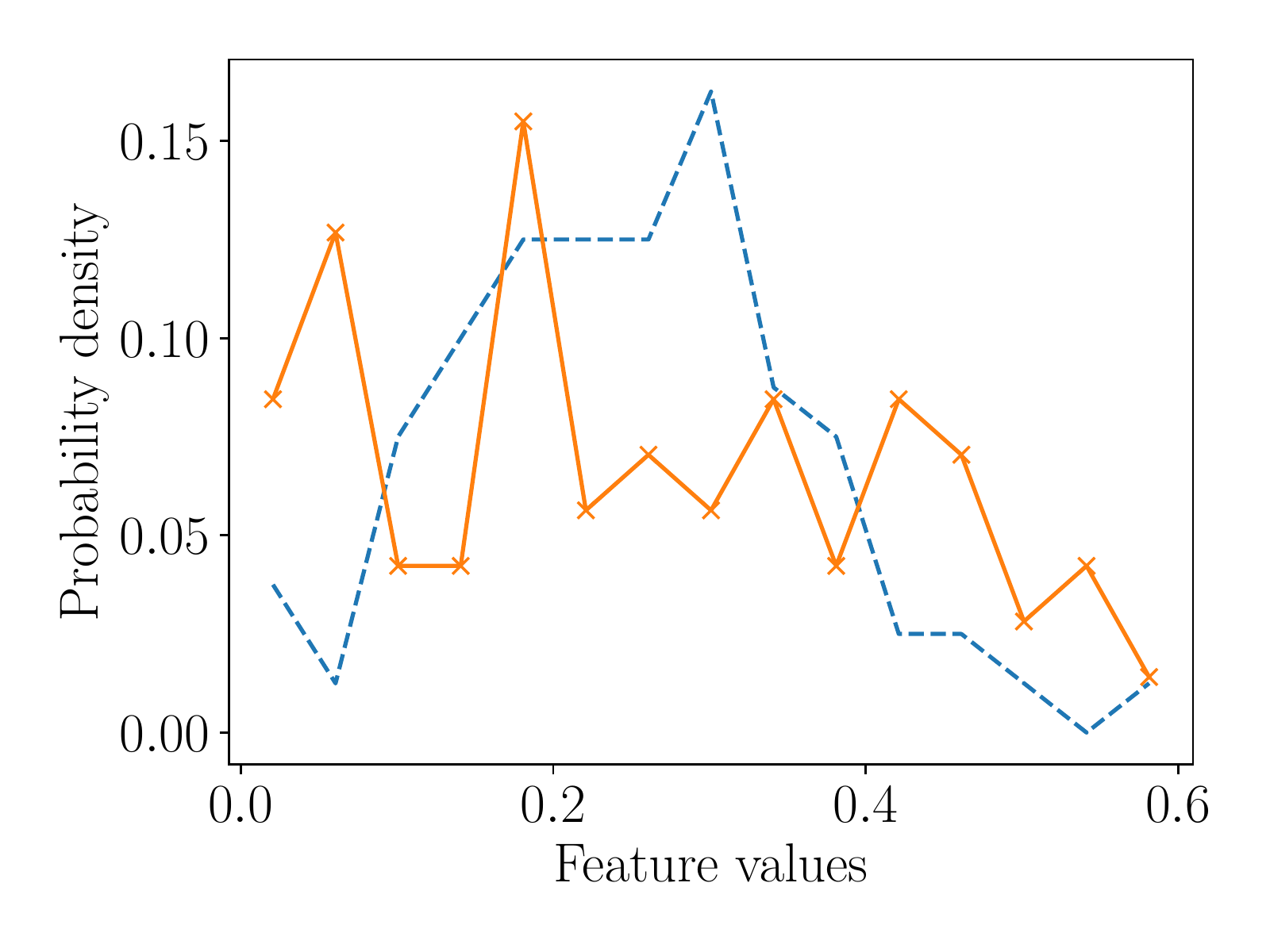}}\\
                
                \scalebox{0.28}{\includegraphics{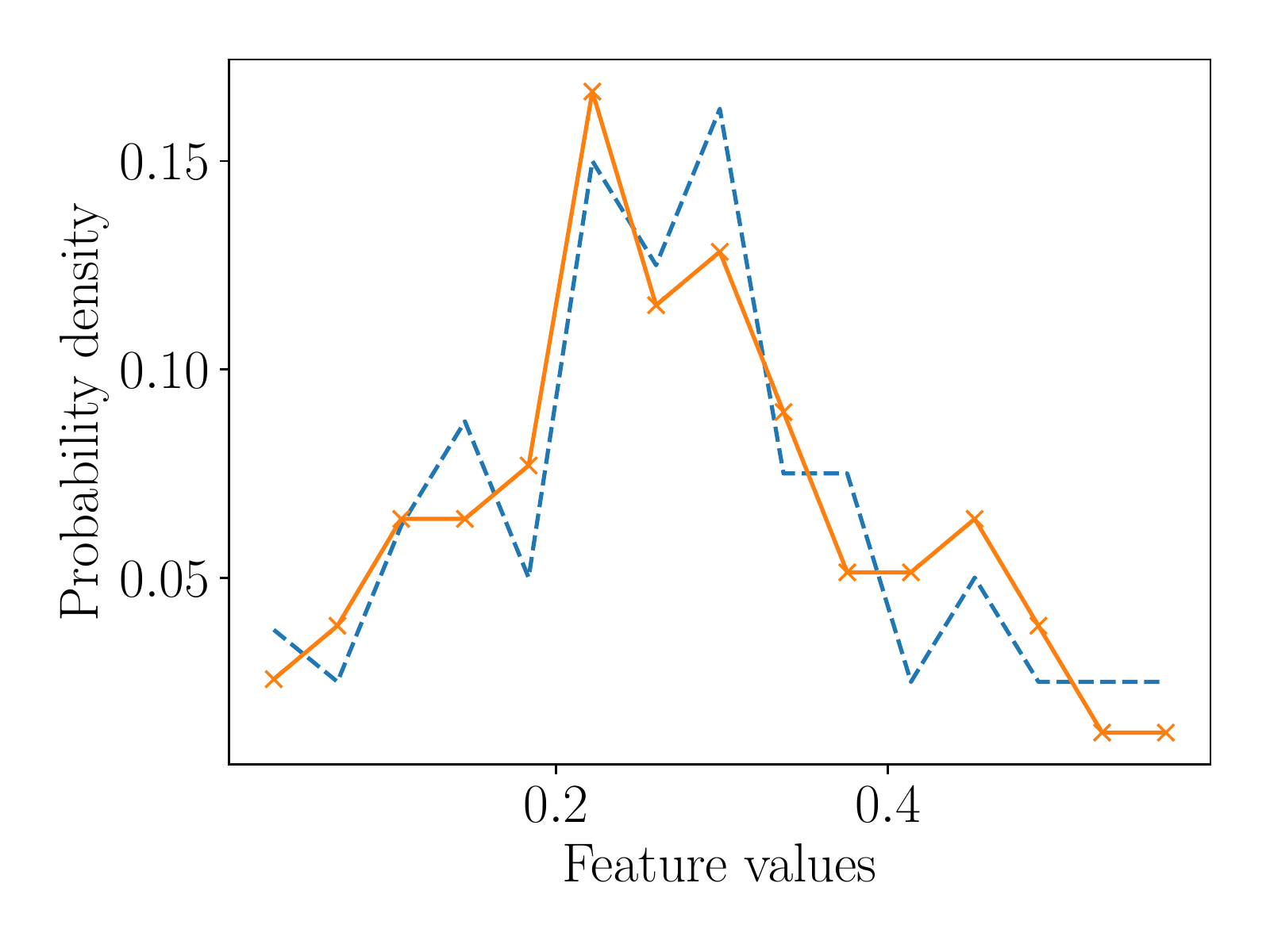}} & \scalebox{0.28}{\includegraphics{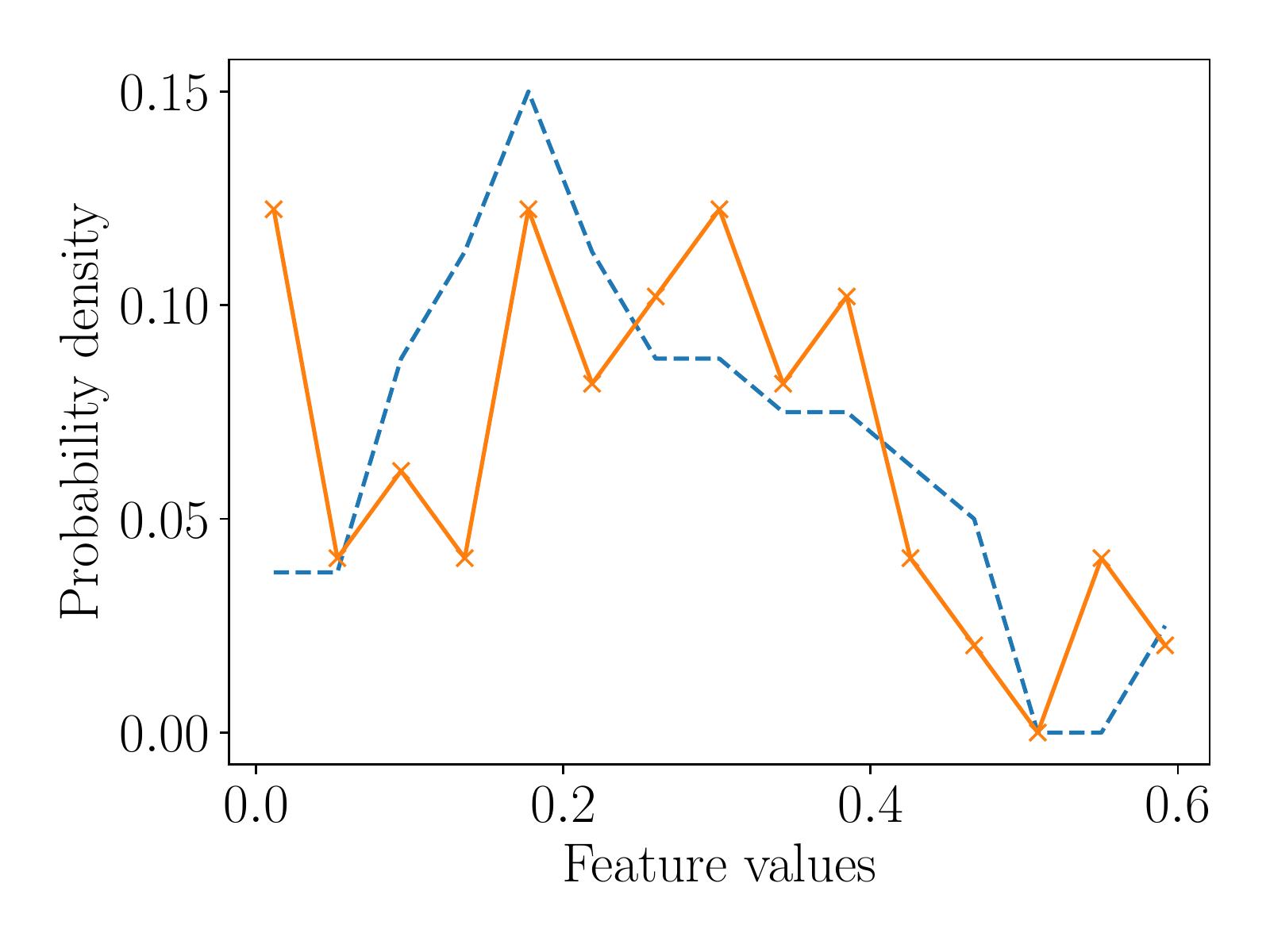}}\\

                \scalebox{0.28}{\includegraphics{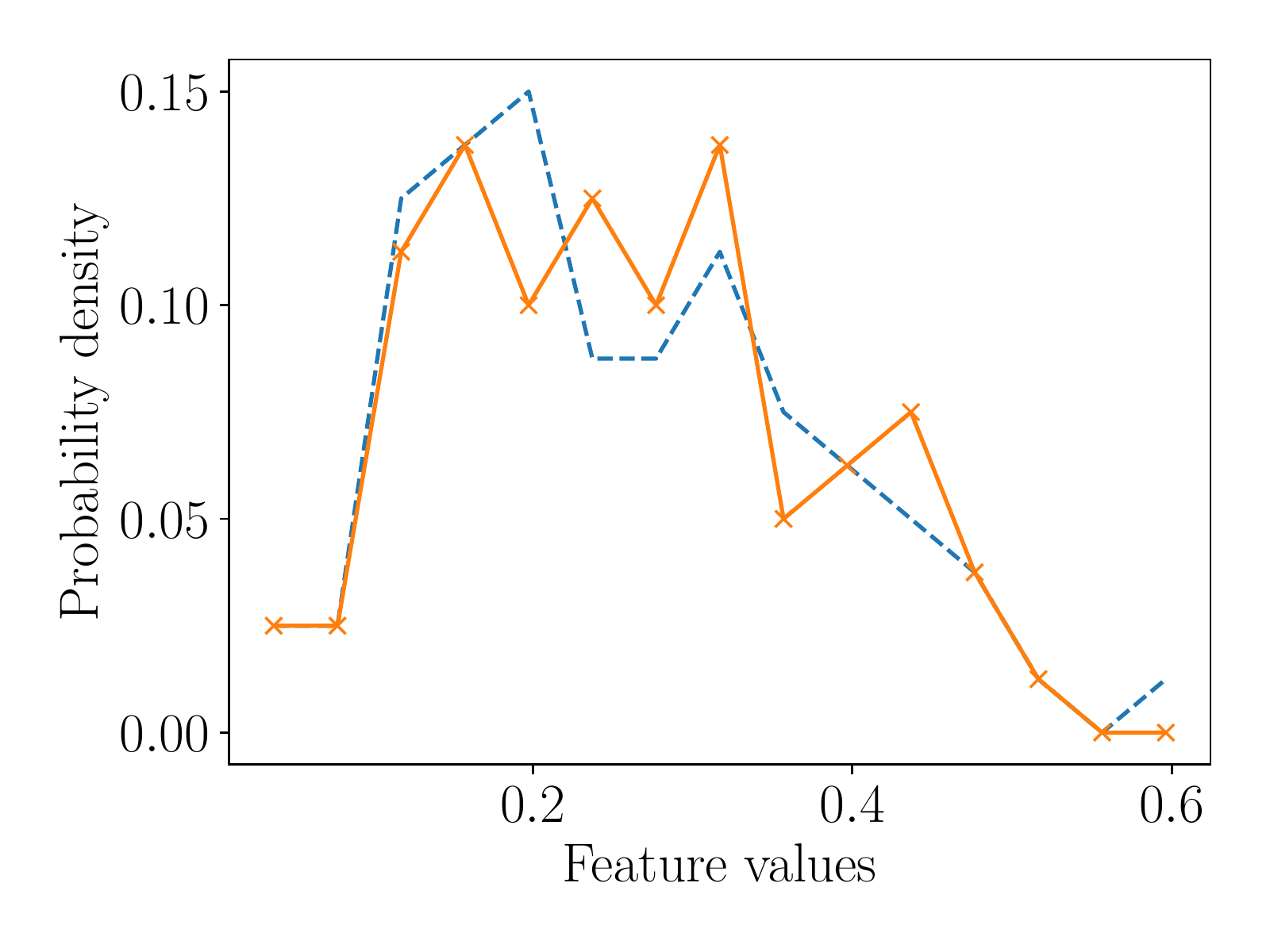}} & \scalebox{0.28}{\includegraphics{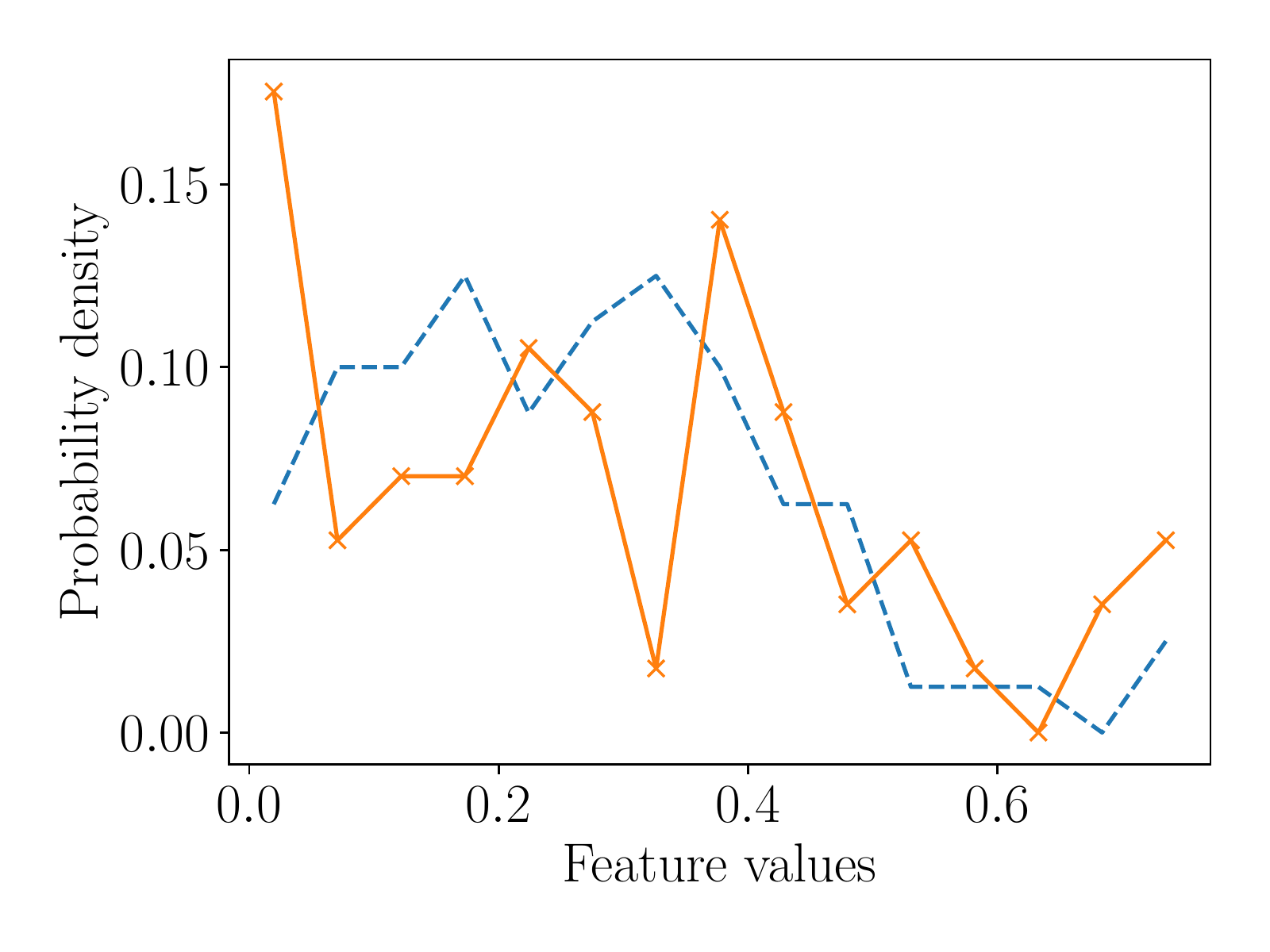}}\\

                \scalebox{0.28}{\includegraphics{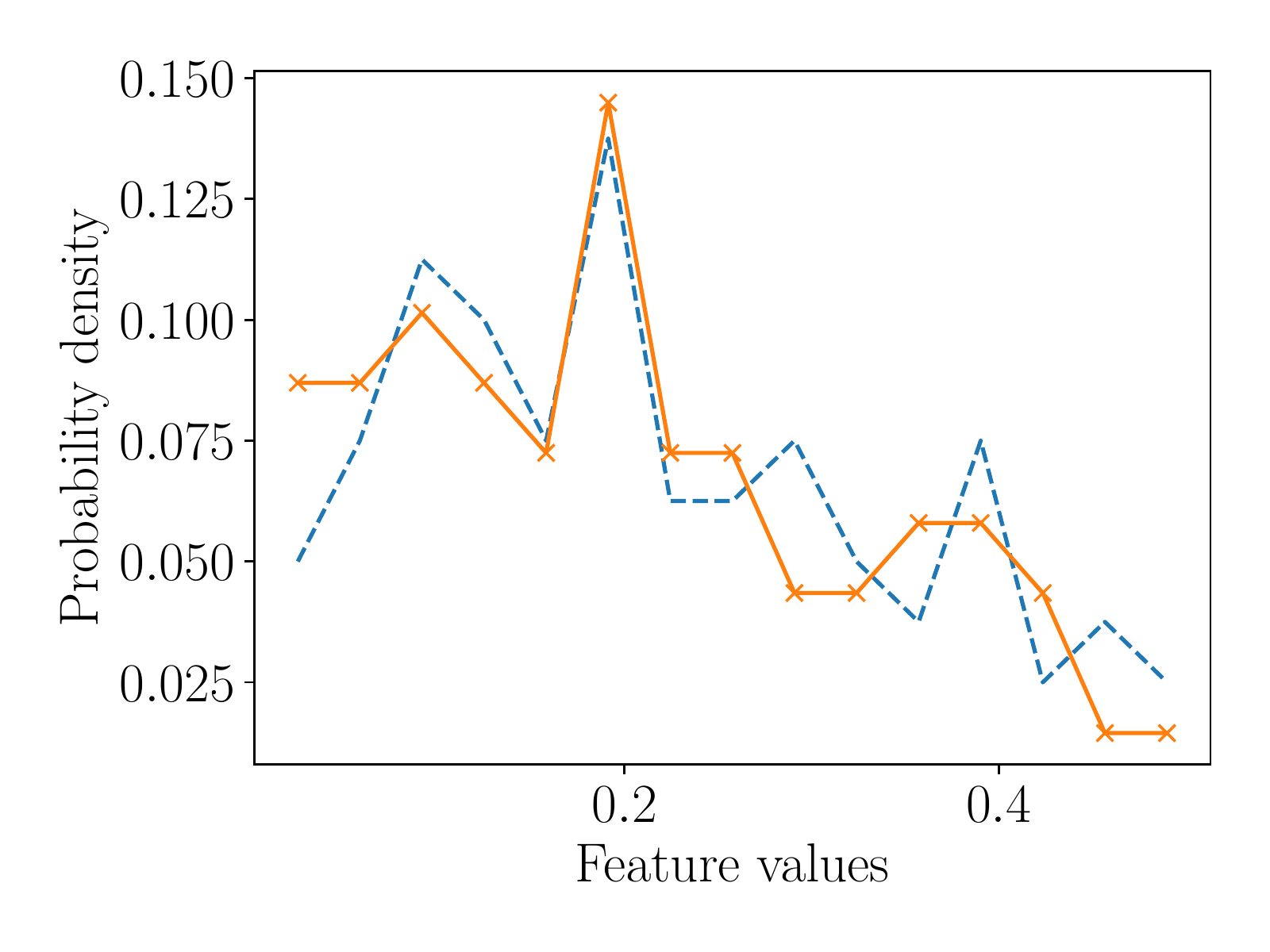}} & \scalebox{0.28}{\includegraphics{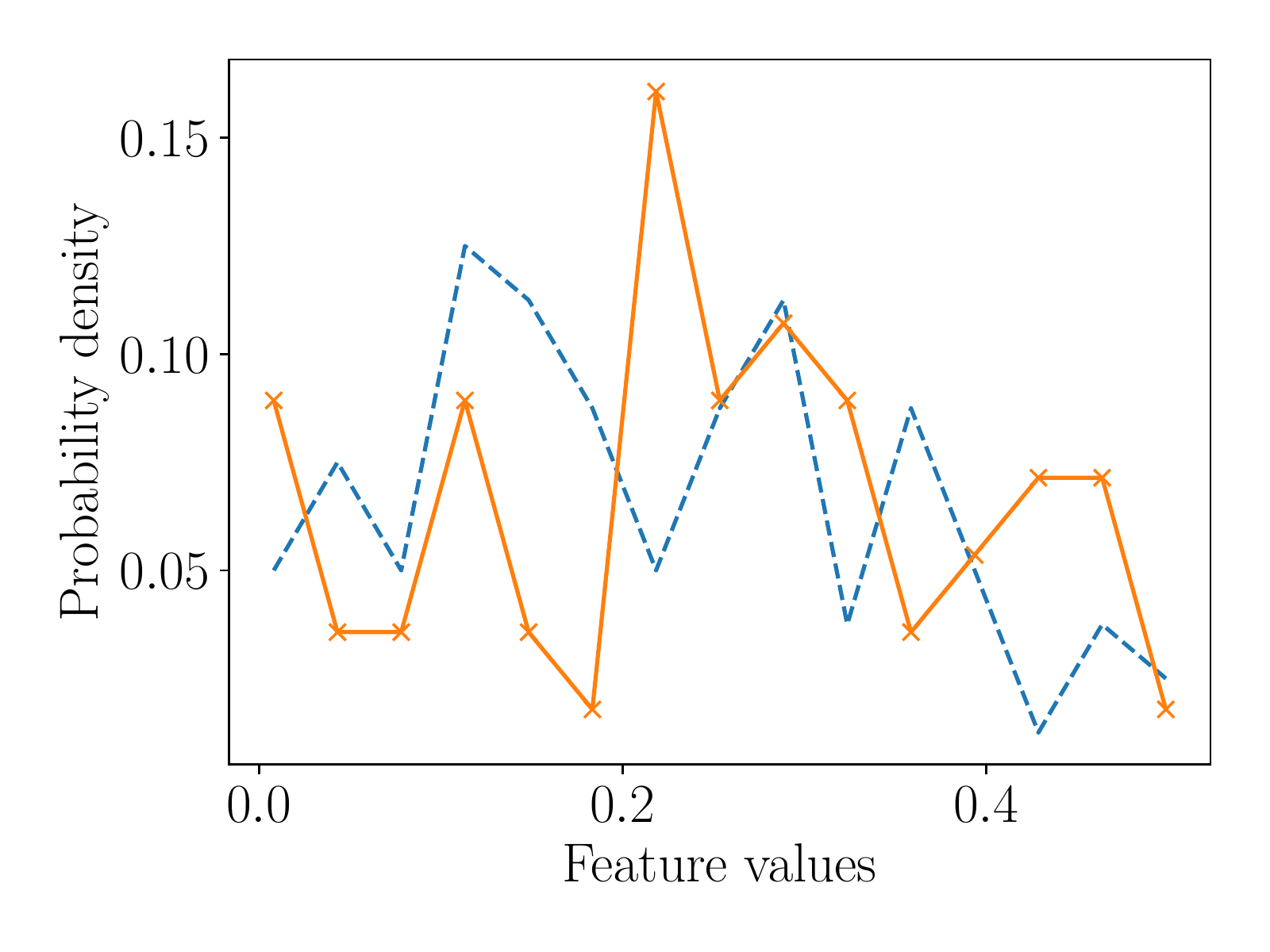}}
                
            \end{tabular}
        }
        \label{tab:Feature_plots}
        \caption{\ Feature distribution for a model trained with MNIST labelled data (orange and continuos in both plots), and ImageNet and \gls{SVHN} unlabelled data (left and right column, respectively,  blue dashed line in both). For each plot a different dataset partition was used.  From top to bottom, for each row: feature 372, 159, 7, 420, 82 and 491. }
    \end{table}

\end{document}